\documentclass[format=acmsmall, review=false, screen=true]{acmart}

\usepackage{hyperref}       
\usepackage{url}            
\usepackage{booktabs}       
\usepackage{amsfonts}       
\usepackage{nicefrac}       
\usepackage{microtype}      
\usepackage{ amsmath, mathtools}
\usepackage{algorithm,algorithmic,dsfont, subfig}
\usepackage{float, soul}

\setcopyright{acmcopyright}
\acmJournal{POMACS}
\acmYear{2017} 
\acmVolume{1}
\acmNumber{2} 
\acmArticle{32} 
\acmMonth{12} 
\acmPrice{15.00}
\acmDOI{10.1145/3154490}

\received{August 2017}
\received[revised]{October 2017}
\received[accepted]{December 2017}

\newcommand{\mathbbm}[1]{\mathds{#1}}

\newcommand{\cN}{\mathcal{N}}

\newcommand{\EE}{\mathbbm{E}}
\newcommand{\NN}{\mathbbm{N}}
\newcommand{\PP}{\mathbbm{P}}

\newcommand{\indic}{\mathbbm{1} }

\newcommand{\ep}{\hfill $\Box$}

\newcommand{\eqs}[1]{ \begin{equation*} #1  \end{equation*}}

\newcommand{\el}{\end{flushleft}}
\newcommand{\bl}{\begin{flushleft}}

\newcommand{\argmax}{\operatornamewithlimits{argmax}}

\renewcommand{\textrightarrow}{$\rightarrow$}

\newcommand{\separator}{
  \begin{center}
    \rule{\columnwidth}{0.3mm}
  \end{center}
}

\begin{document}

\title{Online Learning of Optimally Diverse Rankings}

\author{Stefan Magureanu}
\affiliation{%
  \institution{KTH Royal Institute of Technology}
  \city{Stockholm}
  \country{Sweden}
}
\author{Alexandre Proutiere}
\affiliation{%
  \institution{KTH Royal Institute of Technology}
  \city{Stockholm}
  \country{Sweden}
}
\author{Marcus Isaksson}
\affiliation{%
  \institution{Spotify AB}
  \city{Stockholm}
  \country{Sweden}
}
\author{Boxun Zhang}
\affiliation{%
  \institution{Spotify AB}
  \city{Stockholm}
  \country{Sweden}
}
\begin{abstract}
Search engines answer users' queries by listing relevant items (e.g. documents, songs, products, web pages, ...). These engines rely on algorithms that learn to rank items so as to present an ordered list maximizing the probability that it contains relevant item. The main challenge in the design of learning-to-rank algorithms stems from the fact that queries often have different meanings for different users. In absence of any contextual information about the query, one often has to adhere to the {\it diversity} principle, i.e., to return a list covering the various possible topics or meanings of the query. To formalize this learning-to-rank problem, we propose a natural model where (i) items are categorized into topics, (ii) users find items relevant only if they match the topic of their query, and (iii) the engine is not aware of the topic of an arriving query, nor of the frequency at which queries related to various topics arrive, nor of the topic-dependent click-through-rates of the items. For this problem, we devise LDR (Learning Diverse Rankings), an algorithm that efficiently learns the optimal list based on users' feedback only. We show that after $T$ queries, the regret of LDR scales as $O((N-L)\log(T))$ where $N$ is the number of all items. We further establish that this scaling cannot be improved, i.e., LDR is order optimal. Finally, using numerical experiments on both artificial and real-world data, we illustrate the superiority of LDR compared to existing learning-to-rank algorithms.

\end{abstract}

\ccsdesc[500]{Computing methodologies~Sequential decision making}
\ccsdesc[300]{Information systems~Content ranking}
\ccsdesc[100]{Computing methodologies~Online learning settings}

\keywords{Learning to rank; multi-armed bandits; online learning; diversity}
\maketitle

\section{Introduction}

Search engines have become a critical component of many online services. They answer users' queries by listing relevant documents available online or in the catalog of available products. These engines rely on algorithms that learn to rank items (e.g. documents, songs, products, web pages, ...) so as to present an ordered list maximizing users' satisfaction, i.e., maximizing the probability that there exists a relevant item in the list. One of the main challenges in the design of learning-to-rank algorithms stems from the fact that queries often have different meanings for different users. For example, the meaning of "happy music" may significantly differ across users, the query "jaguar" can be related to cars, the animal, a sport team, etc. In absence of any contextual information about the query, one often has to adhere to the {\it diversity} principle, i.e., to return a list covering the various possible topics or meanings of the query. Ideally, one would wish to learn the list having maximal click-through-rate (i.e., the probability that one item in the list is relevant), but the latter clearly depends on both the unknown frequencies of queries related to the various possible topics, and the unknown topic-dependent click-through-rates of all possible items. Unfortunately, even when all the aforementioned parameters are known, identifying this optimal list is often viewed as a submodular maximization problem, and without specific structural assumptions, it is NP-hard. 

In this paper, we consider the online learning-to-rank problem where the optimal list should be learned in an online manner through users' feedback only. In this problem, the search engine sequentially receives the same query from users interested in various topics. It then returns an ordered list of $L$ items chosen out of $N$ possible items. The user parses the list in order, and clicks on the first item she judges relevant. The engine observes where a click occurred, if any, and refines its displayed list for the next user accordingly. This model of user behavior is known in the literature as the \emph{cascading-click} model. The relevance of the cascading-click model is showcased in \cite{craswell2008experimental}, making it one of the most popular single-click model of user behavior in the learning to rank literature (\cite{schuth2016multileave}, \cite{Combes2015LearningToRank},\cite{kveton2015cascading}, \cite{schuth2013lerot} etc.). We assume that the unknown topic of a user's query is drawn in an i.i.d. manner from a distribution $\phi=(\phi_1,\ldots,\phi_M)$ over the $M$ possible topics. This distribution is also unknown. The click-through-rate\footnote{The probability that the user clicks on the item if inspected in the displayed list.} of item $k$ depends on the topic of the query, and is equal to $\theta_{km}$ for queries of topic $m$. $\theta=(\theta_{km})_{k,m}$ is also initially unknown. The objective is to devise an algorithm sequentially selecting lists, depending on past displayed lists and corresponding feedback, and maximizing the cumulative number of clicks over a fixed but large number of successive queries. Equivalently we look for an algorithm minimizing regret defined as the difference between the average cumulative number of clicks under the optimal list and that achieved under the algorithm. As stated above, even if $\phi$ and $\theta$ were known, identifying the optimal list is NP-hard in general, and as a consequence, so far, the performance guarantees of existing algorithms for this problem are weak: e.g., \cite{radlinski2007active,kohli2013fast,rahman2015fast} propose algorithms whose (prohibitive) regret upper bound scales as $(1-1/e)T+O(NL\log(T))$ after $T$ queries. To circumvent this difficulty, we make the following reasonable structural assumptions: 

\noindent \textbf{Assumption 1)} The set $\cN$ of items is partitioned into $M$ non-overlapping subsets $\cN_1,\ldots,\cN_M$ where each subset corresponds to items related to a particular topic. This partition is assumed to be known, which essentially means that items have been categorized into topics using previous observations, and available meta-data.

\noindent \textbf{Assumption 2)} A user interested in a topic $m$ is very unlikely to click on an item related to other topics, i.e., $\theta_{km}\approx 0$ if $k\notin \cN_m$.

These assumptions are justified in settings where \emph{diversity} is required, i.e., when optimal lists contain items of different topics. If the rewards are similar across topics, there is no real motivation for even considering topics in the model - and such models have been investigated in prior work, e.g. \cite{Combes2015LearningToRank}, \cite{kveton2015cascading}. Presenting diverse lists is of great importance when there exist negative correlations between rewards of items across topics, which is often the case when systems must account for unobserved contextual information. 
Under the above assumption on $\theta$, we could characterize the optimal list using a simple greedy procedure if $\theta$ and $\phi$ were known. Now in this paper, if $\theta$ and $\phi$ are unknown, we devise an algorithm with low regret, namely scaling at most as $O((N-L)\log(T))$, which is provably order-optimal. 

In spite of our simplifying assumptions, the problem inherits most of the challenges of the problem in the general setting: (i) the decision space is very large (there are $N!/(N-L)!$ possible lists), and the user feedback is partial (e.g. we cannot infer the topic of the query if the user does not click on any item). (ii) The sequential list selection algorithm should identify the list with the \emph{optimal level of diversity} among topics (this level is dictated by the initially unknown values of $\phi$ and $\theta$). (iii) Generally, building unbiased estimates of the click-through-rates $\theta$ and of the distribution $\phi$ would require to explore a large number of sub-optimal lists. In fact, we may identify the optimal list without identifying the parameters $\theta$ and $\phi$ individually, but coming up with the most efficient exploration procedure towards this aim is challenging.

As any other online learning algorithm, LDR (Learning Diverse Ranking), our proposed algorithm, carefully balances exploration and exploitation. Its novelty however lies in the fact that it relies on two types of exploration procedure: a first procedure meant to rank all items, and a second aiming at ranking items related to the same topic. We believe that this {\it double} exploration provides an elegant and efficient way to quickly identify the optimal list, without actually estimating the $\theta_{km}$'s and the $\phi_m$'s individually. The novelty of the proof consists in showing the convergence in finite time of the indexes used in the algorithm to proper confidence bounds on relevant quantities, despite their complex appearance. We show that LDR is order-optimal and that it outperforms existing algorithm on artificial and real-world data (we tested the algorithm on data provided by Spotify, one of the most popular music streaming services.

\medskip
\noindent
{\bf Paper organization and contributions.} The next subsection presents the related work. In Section \ref{sec:model}, we present our model in more detail. In Section \ref{sec:algo}, we present the LDR algorithm and derive upper bound of its regret. This bound scales as $O((N-L)\log(T))$. A tight regret lower bound is derived in Section \ref{sec:low}. More precisely, we show that under any algorithm, the regret should scale at least as $\Theta((N-L)\log(T))$ after $T$ queries. Finally, we conduct numerical experiments illustrating the superiority of our algorithm over existing algorithms on artificial and real-world data.

\subsection{Related Work}

We present here three classes of online learning or {\it bandit} problems that are similar to ours, and the corresponding existing results. First, we look at learning to rank problems where the items are not classified into fixed topics (i.e. the mapping of items to topics is hidden and arbitrary from round to round, thus removing the negative correlation structure and the purpose of \emph{diversity}). Second, we look at contextual combinatorial bandits where items are classified into topics, but the classification is not revealed. Third, we look at results concerning combinatorial bandits with cascading feedback. We conclude this section by highlighting the novelty of our model and results.

\medskip
\noindent
{\bf Bandits with Unpartitioned Items.} In \cite{radlinski2008learning} at each round, the user is assigned an arbitrary (and hidden) set of relevant items. In contrast, in our case, the partition of relevant items is fixed, and while the decision maker is aware of the partition, the topic of interest of the user's query is hidden and stochastic. We believe this structuring of items into topics is more realistic than that considered in \cite{radlinski2008learning}, where the collection of items relevant to the users is considered adversarial in nature. A more generic, still adversarial, setting is also considered in \cite{streeter2009online} where the authors consider the reward at each round to be a hidden submodular function of the displayed set. The setting in \cite{radlinski2008learning} is further generalized in \cite{slivkins2013ranked} to better account for similarities between rankings, however the authors only consider Lipschitz continuity and not negative correlation (a user querying "jaguar" will be either interested in the animal or the car, not both).

Another similar setting is studied in \cite{yue2011linear}. Here the system is assumed to interact with only one user. At each round, new items arrive and are represented by a set of coverage functions that indicate the relevance of the item to each topic, for any combination of items placed ahead of it. These functions are revealed to the decision maker at the beginning of each round. The decision maker must then present $L$ items which are scanned from top to bottom by the user who clicks on an item with a probability dictated by the coverage functions of the item and an unknown fixed feature vector of the user $w^*$. The system receives feedback for every item presented and the next round begins. Unlike in our setting, here, at every round, new items arrive and feedback does not respect the cascading model. Also, the contextual information (here, the coverage functions) is always revealed, and furthermore, available to the decision maker before they need to present an action. We believe our setting presents more practical relevance as the coverage functions are hard to obtain in real life, whereas our setting imposes a need for diversity in a very natural way. The setting of \cite{yue2011linear} is further extended in \cite{yu2016linear} to include knapsack constraints.

\medskip
\noindent
{\bf Contextual Combinatorial Bandits.} Most related to our setting, in \cite{kohli2013fast}, the authors assume queries have i.i.d. relevance vectors (in round $t$, this binary vector $X^t\in \{0,1\}^N$ determines whether items are relevant or not), drawn from an unknown distribution $D$. This model can account for negative correlations among item relevances and for a classification of items and queries into topics, and hence would push \emph{diversity} in the selected lists. However, this classification is not revealed to the decision maker, as in our case, and therefore the decision maker cannot take advantage of the negative correlations among items of different topics. Other similar settings are investigated in \cite{rahman2015fast},\cite{qin2014contextual} and \cite{li2016contextual}. Critically, both \cite{kohli2013fast} and \cite{rahman2015fast} consider that the decision maker may observe the relevance of every displayed item at the end of the round, rather than just the first scanned relevant item. The authors propose the use of a \emph{per slot} UCB approach, which considers an independent bandit problem in each slot and uses an \emph{off-the-shelf} algorithm (i.e. UCB, exp3, $\epsilon$-greedy) in each slot. For this class of algorithms, they obtain regret guarantees scaling as $(1-1/e)T+O(LN\log(T))$, whereas we show it is possible to obtain regret scaling as $O((N-L)\log(T))$ - despite only observing the position of the first relevant item and not that of all relevant items. Our regret guarantees are missing the $(1-1/e)T$ term since in our setting the optimal list can be computed in polynomial time whereas in \cite{kohli2013fast}, the regret is computed relative to the best approximation of the offline optimal list that can be computed in polynomial time. In \cite{li2016contextual} and \cite{qin2014contextual}, the context of each user is revealed and hence, the need for \emph{diversity} is removed as the decision maker does not need to present a mixed list of items, each performing well under different contexts. Additionally, the estimation of rewards becomes substantially easier when the context is revealed.

\medskip
\noindent
{\bf Contextual Cascading Bandits.} We differentiate our work from that in \cite{Combes2015LearningToRank} (and other articles sharing their setting, such as \cite{kveton2015cascading}) by not revealing the topic of a query upon arrival, hence requiring to account for the \emph{diversity} of the results. In their work, Combes et al. make a similar assumption as ours on the classification of queries and items into classes and topics, respectively. They consider two cases, one when the topic of interest of the class of users is known, and one when it has to be learned, however, the class of the query is always revealed at the beginning of the round. Our setting is mentioned when discussing the \emph{diversity} principle, but there are no results regarding this scenario. In contrast to the settings considered in \cite{Combes2015LearningToRank}, we consider the decision maker has no information regarding the classes of arriving queries (it is not directly observed, and furthermore, their distribution is unknown). This significantly complicates the task of the learner which must now consider ranked lists containing \emph{diverse} entries. Note that in \cite{Combes2015LearningToRank}, all algorithms commit to presenting items from a single topic, and disregard diversity.

\medskip
\noindent
{\bf Model Novelty.} Our setting can be viewed as a stochastic contextual combinatorial bandit with cascading feedback, where the context (here the topic of the query) is \emph{hidden}, randomly selected at each round and defines the average rewards of various items. Existing work on contextual bandits assume that the context is revealed to the decision maker before they select an arm (here a list of items). The novelty of our model is that the context is \emph{not} revealed, nor its distribution $\phi$. If a user does not click, the context of her query is not revealed, and when she clicks, the context is revealed {\it a posteriori} after the click occured. This complicates the design of algorithms, and in particular, since we allow multiple plays, calls for selecting diverse lists (since we do not know the arriving context or topic, we would like to present relevant items from each topic).

\section{Preliminaries}\label{sec:model}

\subsection{Model}

The set $\mathcal{ N}$ of $N$ items is partitioned into $M$ non-overlapping subsets $\mathcal{N}_1,\dots,\mathcal{N}_M$, each containing items related to a given topic. The decision maker is aware of this partition. We define the mapping $h:\mathcal{N}\to \{1,\dots, M\}$ such that for any item $k$, $h(k)$ denotes its topic, i.e. $h(k)=m$ iff $k\in \mathcal{N}_m$. The click-through-rate (CTR) of item $k$ depends on the topic of the query: for a query related to topic $m$, the user finds $k$ relevant with probability $\theta_{km}$. $\theta=(\theta_{km})_{k,m}$ is unknown, but we assume that $\theta_{km}=0$ whenever $k\notin \mathcal{ N}_m$ (a user interested in topic $m$ finds items not related to $m$ irrelevant). Queries arrive at the decision maker sequentially, and the topic of the $n$-th query, denoted by $m(n)$, is unknown. $(m(n))_{n\ge 1}$ is an i.i.d sequence of r.v. with values in $\mathcal{ M}=\{1,\ldots,M\}$, and with distribution $\phi=(\phi_1,\ldots,\phi_M)$, also unknown to the decision maker. Note that the notion of {\it query} is loosely defined here: a query may for example corresponds to text strings containing a particular set of keywords (e.g. Christmas). We look at instants or rounds where the engine receives the same query. 

After receiving the query in the $n$-th round, the decision maker returns an ordered list $u(n)=(u_1(n),\ldots,u_L(n)) \in \mathcal{ N}^{L}$ of $L$ items ($L$ is typically much smaller than $N$). The user then scans the items in the list in order, and clicks on the first relevant item, if any. If the user clicks on an item, the decision maker observes the slot or position of the corresponding item in the list, and gets a unit reward\footnote{As in \cite{Combes2015LearningToRank}, we can generalize our model and results to the case where the reward depends on the position of the first relevant item in the list.}. The decision maker gets no reward if the user does not click on any item. For the $n$-th query, a binary random vector $X(n)=\{X_l(n): l=1,\dots,L\}$ indicates whether the various items in the displayed list are relevant, i.e., $\PP[X_l(n) = 1| u(n) = u, m(n) = m] = \theta_{u_lm}$. Given the sequence of queries and displayed lists, the r.v. $(X_l(n))_{n\ge 1,l}$ are independent. The average reward of a list $u$ is then: 
$$
\mu_{\phi,\theta}(u)= \sum_{m=1}^M \phi_m\sum_{l=1}^L \theta_{u_lm}\prod\limits_{i=1}^{l-1}(1-\theta_{u_im}).$$

Throughout this paper we will use the following shorthand notation: $\mu_{\theta,\phi}(u) = \mu(u)$. We also denote by $u^\star=u^\star(\phi,\theta)=\arg\max_u \mu_{\theta,\phi}(u)$ the optimal list (assumed to be unique for simplicity), and we assume without loss of generality that
$u^\star=\{1,\ldots,L\}$. A sequential decision policy $\pi$ selects lists depending on the previous selected lists and the corresponding users' feedback, and we denote by $u^\pi(n)$ the list chosen under $\pi$ for the $n$-th query. We denote by $\Pi$ the set of such policies. The problem is to identify $\pi\in \Pi$ minimizing its regret $R^\pi(T)$ after $T$ queries, where:
$$
R^\pi(T) = T\mu(u^\star) - \EE[\sum\limits_{n=1}^T \mu(u^{\pi}(n))].
$$

\subsection{Computing the Optimal List \texorpdfstring{$u^\star$}{Lg} }\label{sec:optimalPolicy}

We establish that when the parameters $\phi$ and $\theta$ are known, we can identify the optimal list $u^\star$ using a low complexity recursive greedy procedure. This is possible only thanks to the structural assumption made on $\theta$ (in absence of such assumption, computing $u^\star$ is NP-hard as stated earlier). We first introduce the  \emph{success rate} $\nu(l | u)$ at position $l$ in an ordered list $u$ as
$$ \nu(l|u) = \sum_{m=1}^M \phi_m \theta_{u_lm} \prod\limits_{s=1}^{l-1} (1-\theta_{u_sm}).$$ 
The success rate $ \nu(l|u)$ is the probability that the item in position $l$ is clicked if the displayed list is $u$. It does not depend on items listed below $l$ in $u$ or on the order of items listed ahead of $l$ in $u$.

We prove that the following recursive greedy procedure outputs $u^\star$. In what follows, we denote by $u^{\star l}$ the list of length $l$ with maximal average reward, i.e., $u^{\star l}$ maximizes among all list $u$ of length $l$: $\sum_{m=1}^M \phi_m\sum_{g=1}^l \theta_{u_gm}\prod\limits_{i=1}^{g-1}(1-\theta_{u_im})$. The recursive procedure sequentially constructs lists $u^{[1]},\ldots,u^{[L]}$ of increasing length. We will establish that $u^{[L]}=u^\star$.

\begin{enumerate}
\item Set $u^{[1]} = \{ k_1\}$ where $k_1=\arg\max_{k\in \mathcal{ N}} \sum_{m\in \mathcal{ M}} \phi_m\theta_{km}$.
\item For $l=2$ to $l=L-1$, given $u^{[l]}=\{k_1,\ldots, k_l\}$, denote by $U(u^{[l]})$ the set of lists of length $l+1$ of the form $\{k_1,\ldots, k_l,k\}$ for $k\in \mathcal{ N}$. Then $u^{[l+1]}=\arg\max_{u\in U(u^{[l]})} \nu( l+1|u)$.
\end{enumerate}

\begin{proposition}\label{lem1}
The above greedy procedure returns $u^\star$, namely $u^{[L]}=u^\star$.
\end{proposition}

\noindent
{\bf Proof.} Denote by $[u | k]$ the list obtained by appending the item $k$ at the end of list $u$. Define  $k(l) = \argmax_{k\notin u^{[l]}} \nu(l+1,[u^{[l]}|k])$, the item offering the highest \emph{success rate} when appended at the end of the list $u^{[l]}$. The above procedure appends the item $k(l)$ to $u^{[l]}$. We show by induction on $l$ that $u^{[l]}=u^{\star l}$ for all $l=1,\ldots,L$. 

The result holds for $l=1$ by definition of $k_1$. Assume that the result holds up to $l-1$, i.e., that $u^{[j]}=u^{\star (j)}$ for all $j\le l-1$. We show that the list $u'=[u^{\star (l-1)}|k(l-1)]$ is optimal, i.e., $u'=u^{\star l}$. Since $u^{[l-1]}=u^{\star (l-1)}$, this implies that $u^{[l]}=u^{\star l}$. In view of our assumption on $\theta$, in the optimal list of length $l$, the items related to topic $m$ are those with highest CTRs  $\theta_{km}$. Also note that the order in which items are placed in a list does not affect its reward. 

Note that $\mu(u^{\star l})\geq \mu(u')$ implies the following relation between the probabilities of each list not containing any interesting items:
$$
\sum\limits_{m=1}^M \phi_m \prod\limits_{s=1}^l (1-\theta_{u^{\star l}_sm}) \leq \sum\limits_{m=1}^M \phi_m \prod\limits_{s=1}^l (1-\theta_{u'_sm} )
$$

We proceed by contradiction: assume that $\mu(u')<\mu(u^{\star l})$. We first establish the following fact.\\
{\it Fact 1.} We cannot have: $\exists k, k'\in\mathcal{N}$ such that $h(k') = h(k)$ and $k\in u^{\star l-1}\setminus u^{\star l}$ and $k' \in u^{\star l}\setminus u^{\star l-1}$. \\
Indeed this case corresponds to a scenario when an item from the list $u^{\star l-1}$ is replaced by one of the same topic in list $u^{\star l}$. This is impossible since $k' \notin u^{\star l-1}$ and $k\in u^{\star l-1}$ implies $\theta_{k'm}<\theta_{km}$ for all $m\in\mathcal{M}$.

Now from Fact 1, and $\mu(u')<\mu(u^{\star l})$ ($u'$ contains all items of $u^{\star (l-1)}$), there exists an item $k$ such that $k\in u^{\star (l-1)}\setminus u^{\star l}$, and an item $k'$ such that $k'\in u^{\star l}\setminus u^{\star (l-1)}$ with $h(k')\neq h(k)$. Let $v$ a reordering of $u^{\star l}$ such that $k'$ is at the end of $v$ (at the $l$-th position). Further define $w$ the list obtained from $v$ by replacing $k'$ by $k$. Now if we show that (a) $\mu(w)\ge \mu(v)$, we can repeat the argument (if there are several couples of items $k\in u^{\star (l-1)}\setminus u^{\star l}$, $k'\in u^{\star l}\setminus u^{\star (l-1)}$ with $h(k')\neq h(k)$) until $u^{\star (l-1)}\subset w$ in which case (a) yields a contradiction. Indeed, since the order of items does not matter, we have (b) $\mu(v) = \mu(u^{\star l})$. Note also that $w$ contains $u^{\star (l-1)}$, and hence by construction of $u'$, (c) $\mu(u')\ge \mu(w)$. Combining (a)-(b)-(c), we get $\mu(u')\ge \mu(u^{\star l})$, a contradiction.

We conclude the proof by establishing that $\mu(w)\ge \mu(v)$. For convenience, we first restate the definition of the \emph{success rate} of a slot $l$ in a list $u$, $\nu(l|u)$:
\begin{align*}
\nu(l|u) & = \sum_{m=1}^M \phi_m \theta_{u_lm} \prod\limits_{s=1}^{l-1} (1-\theta_{u_sm})\\
& = \phi_{h(u_l)} \theta_{u_lh(u_l)} \prod\limits_{s=1}^{l-1} (1-\theta_{u_sh(u_l)})
\end{align*}

Observe that all items of topic $h(k')$ in $u^{\star (l-2)}$ must be in $v$. This is due to Fact 1: since $k'\in u^{\star l}\setminus u^{\star (l-1)}$, all items of topic $h(k')$ in $u^{\star (l-1)}$ are in $u^{\star l}$ and hence in $v$. But by our induction hypothesis, $u^{\star (l-2)}\subset  u^{\star (l-1)}$, and thus indeed all items of topic $h(k')$ in $u^{\star (l-2)}$ are in $v$. From this, we deduce that:
\begin{equation}\label{ineq:1}
\nu(l-1| [u^{\star (l-2)}|k']) \geq \nu(l|v).
\end{equation} 
By construction of $u^{\star (l-1)}$, we have: 
\begin{equation}\label{ineq:2}
\nu(l-1| u^{\star (l-1)}) \geq \nu(l-1| [u^{\star (l-2)}|k']).
\end{equation}
Note also that by construction of $u^{\star (l-1)}$ and due to the induction hypothesis, for all $j\leq l-1$:
\begin{equation}\label{eq:auxineq}
\nu(j|u^{\star (l-1)})\ge\nu(l-1| u^{\star (l-1)}).
\end{equation}
In view of Fact 1, no additional items of topic $h(k)$ are present in $u^{\star l}$ apart from the ones in $u^{\star l-1}$. Hence we have:
$$
\nu(l|w) \geq \phi_{h(k)} \theta_{kh(k)} \prod\limits_{s=1}^{l-1} (1-\theta_{u^{\star (l-1)}_sh(k)}\mathbbm{1}[u^{\star (l-1)}_s \neq k]). 
$$
Now define $d = \max\{a\leq l-1: h(u^{\star (l-1)}_a) = h(k)\}$ the last slot containing an item of topic $h(k)$ in $u^{\star (l-1)}$. We have:
$$
\nu(d|u^{\star (l-1)}) = \phi_{h(k)} \theta_{u^{\star (l-1)}_d h(k)} \prod\limits_{s=1}^{l-1} (1-\theta_{u^{\star (l-1)}_sh(k)}\mathbbm{1}[s \neq d]).
$$
By construction of $u^{\star (l-1)}$, $\theta_{kh(k)}\geq \theta_{u^{\star (l-1)}_d h(k)}$ and thus:
$$
\frac{\nu(l|w)}{\nu(d|u^{\star l-1})} \geq \frac{\theta_{kh(k)}(1-\theta_{u^{\star l-1}_d h(k)})}{\theta_{u^{\star l-1}_d h(k)} ( 1 - \theta_{kh(k)})} \geq 1.
$$
We conclude that:
\begin{equation}\label{ineq:3}
\nu(l|w) \geq \nu(d|u^{\star l-1})\geq \nu(l-1|u^{\star l-1}),
\end{equation}
where the last inequality is deduced from \eqref{eq:auxineq}. Hence, from \eqref{ineq:1}, \eqref{ineq:2} and \eqref{ineq:3} we have: 
\begin{equation}\label{eq:last}
\nu(l|v)\leq \nu(l|w).
\end{equation} 
Since $v$ and $w$ only differs in the last slot $l$, from \eqref{eq:last}, we have $\mu(v)\le \mu(w)$, which concludes the proof.\ep

\section{The LDR Algorithm and its Regret}\label{sec:algo}


In this section, we first present the rationale behind the design of the LDR algorithm, and then describe its various components in detail. We finally derive a regret upper bound for LDR, scaling as $O((N-L)\log(T))$. Hence in view of the lower bound presented in the previous section, LDR is order-optimal.

As the recursive construction of the optimal list $u^\star$ suggests, any good sequential list selection policy should get accurate estimates of some success rates $\nu(l | [u^{\star (l-1)},k])$ for all item $k$ as defined in Section \ref{sec:optimalPolicy}. This could involve a rather heavy exploration, and generate too much regret. The design of the LDR algorithm is guided by the principle of {\it parsimonious} exploration: the algorithm maintains a {\it leader}, the list believed to be optimal and denoted by $u^\star(n)$ in round $n$, and when it explores (when the leader is not displayed), the explored list is just a slight modification of the leader. More precisely, LDR explores items only in the first and last slot of lists. This parsimonious exploration would confer to the algorithm a low regret, but only if this exploration is sufficient to learn $u^\star$ quickly. 

\subsection{LDR Double Exploration}

The LDR algorithm is designed so as the {\it leader} $u^\star(n)$, that LDR regularly updates, rapidly converges to the true optimal list $u^\star$. Most of the time, LDR exploits the leader (i.e., LDR plays the leader), and it explores when needed. The exploration rounds and the way the leader is updated in LDR are jointly designed so that $u^\star(n)$ converges to $u^\star$: first there are exploration rounds (type-2 exploration) where LDR replaces the item in the first position of the leader by an item $k$ to estimate the parameter $\phi_{h(k)}\theta_{kh(k)}$. In turn, these type-2 exploration rounds allow us to rank items related to the same topic. Now assume that for a given topic $m$, we know the ranking of items related to $m$. LDR leverages this ranking when updating the leader; and more precisely, it lists items with better ranks first. Under such a construction of the leader, the empirical success rate of an item related to $m$ actually corresponds to the true success rate the item would get if  displayed in the optimal list $u^\star$. Indeed, assume that $k_1,\ldots,k_p$ are items related to topic $m$ and that $k_i$ has higher CTR than $k_{i+1}$ for any $i$. Then knowing this order, the empirical success rate of item $k_j$ when displayed in the leader will converge to $\phi_m\theta_{k_jm}\prod_{i=1}^{j-1}(1-\theta_{k_im})$, which is the actual success rate if $k_j\in u^\star$. As a consequence, LDR can use this empirical success rate to determine whether the item should belong to $u^\star$ (comparing with other items). Specifically, LDR has exploration rounds (type-1 exploration) where the last item of the leader is replaced by an apparently sub-optimal item so as to confirm whether the latter is indeed sub-optimal. 

To summarize, LDR runs two types of exploration: Type-2 exploration, which uses the first slot, determines the order of items within their corresponding topic and Type-1 exploration, which uses the last slot and ensures none of the apparently sub-optimal items can favorably replace the last (and therefore \emph{worst}, due to the greedy construction if the leader) item in the leader.

\medskip
\noindent
{\bf Type-1 Exploration.} LDR explores, in the last slot of lists, items that are not in the leader, but that could advantageously replace the item placed last in the leader (i.e., the weakest item in the leader). The upper confidence bound index for item $k$ is here defined as a classical KL-UCB index (\cite{garivier2011kl}):
$$
d_k(n) = \sup\{x \in (0,1): t_k(n) I(\hat c_k(n), x)\leq f(n)\},
$$
where $f(n)= \log(n)+4\log\log(n)$, $t_k(n)$ is a counter incremented if $k$ is displayed and either (i) the true (i.e. not shuffled) leader is displayed or (ii) LDR performs a Type-1 exploration, and $\hat{c}_k(n)$ denotes the empirical success rate of item $k$ (the number of clicks on item $k$ divided by $t_k(n)$). The second type of exploration, together with the fact that in LDR, the leader is updated using the values of the $\hat{c}_k(n)$'s, ensure that $\hat{c}_k(n)$ converges the desired success rate of $k$. More precisely, when $k\in u^\star$, it converges to the success rate of $k$ when displayed in $u^\star$. When $k\notin u^\star$, it converges to the success rate of $k$ when displayed in the last slot of $u^\star$.

\medskip
\noindent
{\bf Type-2 Exploration.} LDR explores item $k$ in the first position of the list, to get an estimate of $\phi_{h(k)}\theta_{k h(k)}$ and hence to be able to rank items related to the same topic.  
$$
b_k(n) = \sup\{x \in (0,1): \tau_k(n) I(\hat\theta_k(n), x) \leq f(n)\},
$$
where $\tau_k(n)$ denotes the number of times item $k$ has been displayed up to round $n$ in a list where no other item of the same topic was placed before $k$, and $\hat\theta_k(n)$ is its empirical success rate for such events -- hence $\EE[\hat\theta_k(n)] = \phi_{h(k)}\theta_{k h(k)}$. $b_k(n)$ is an upper confidence bound on $\phi_{h(k)}\theta_{k h(k)}$. 
\medskip

{It should be observed that the second exploration procedure is necessary. Indeed assume for example that item $k$ is the only item related to topic $m$ in the leader so that when exploring item $j$ of the same topic (using the first exploration procedure), the observable success rates for $k$ and $j$ are $\phi_m\theta_{km}$ and $\phi_{m}(1-\theta_{km})\theta_{jm}$, respectively. The knowledge of these rates is not enough to determine whether e.g. $\theta_{jm} >\theta_{km}$\footnote{Indeed, the two following sets of parameters lead to the same values for $\phi_m\theta_{km}$ and $\phi_{m}(1-\theta_{km})\theta_{jm}$. Set 1: $\theta_{km}=5/7$, $\theta_{jm}=0.5$, $\phi_m=0.7$, here $\theta_{km}>\theta_{jm}$. Set 2: $\theta_{km}=5/6$, $\theta_{jm}=1$, $\phi_m=0.6$, here $\theta_{km}<\theta_{jm}$.}, i.e., whether $j$ should replace $k$ in the leader. }


\subsection{Detailed Implementation}

Next we describe LDR in more detail; its pseudo-code is provided in Algorithm \ref{algo}. The algorithm splits time into windows of $4$ rounds. At the beginning of each window, LDR computes the list currently believed to be optimal (the \emph{leader}). The first and last slots of this window are reserved for exploitation and the second and third are dedicated to exploration (Type-2 and Type-1, respectively). {Note that the order of the procedures in each round is important for the analysis.} Below we detail how LDR executes each of these functions.

\medskip
\noindent \textbf{Updating the leader $u^\star(n)$.} Every $4$ rounds (i.e., when $n\mod 4 = 0$), the algorithm updates the leader, denoted by $u^\star(n)$. To generate the leader, the algorithm orders the items in descending order of $\hat c_k(n)$ and then substitutes the entries from each topic $m$ with the items $k$, $h(k)=m$, maximizing $\hat\theta_k(n)$. In other words, when computing $u^\star(n)$, we first decide how many items from each topic should be present in the leader, and we then place the best items of each topic in the list. We will show that using this construction, the leader indeed converges the optimal list. 

\medskip
\noindent \textbf{Exploration \& Exploitation.} Denote by $W = n \mod 4$. LDR then proceeds as follows.
\begin{itemize}
\item If $W = 0$, LDR exploits and plays the true leader $u^\star(n)$ (Event 1a), 
\item If $W = 3$, LDR plays a shuffled version of the leader (Event 1b). The goal of this phase is to perform Type-2 exploration for items in the leader and ensure both that the items in all topics are ordered in descending order of their click-through rates and that the estimates $\hat\theta_{k}(n)$ are accurate for all $k\in u^\star(n)$. Note that shuffling $u^\star(n)$ does not influence the expected reward of the round.
\item Otherwise, LDR investigates the opportunities to explore. 
\begin{itemize}
\item If $W = 1$, it first looks whether a Type-2 exploration would be relevant. To this aim, it checks whether there is an item $k$ not in the leader with index $b_k(n)$ greater than $\hat\theta_j(n)$ for some item $j$ in the leader and of the same topic as $k$. Should this case arise (if several items satisfy the exploration condition, one is chosen uniformly at random), it places $k$ in the first position of the leader and explores with the resulting list (Event 2), else it moves to the next step.
\item LDR looks whether a Type-1 exploration would be relevant. To this aim, it checks whether there is an item $k$ not in the leader with index $d_k(n)$ greater than $\hat{c}_{u_L}(n)$, and with different topic $h(k)\neq h(u_L(n))$. Should this case arise (ties are broken uniformly at random if several items satisfy the exploration condition), it places $k$ in the last position of the leader and explores with the resulting list (Event 3), else it plays the leader (Event 4).
\end{itemize}
\end{itemize}
After LDR plays the selected list, it updates all decision variables. Note that the success rates $\hat c_k(n)$ and the corresponding counter $t_k(n)$ are only updated if Events {1a}, {3} or {4} occur. 

\begin{algorithm}[ht!]
\caption{LDR (Learning Diverse Rankings)}\label{algo}
\begin{algorithmic}
\STATE Initialize for all item $k$ : $\hat c_k(0) = 0.5$, $\hat \theta_k(0) = 0.5$, $t_k(0) = 1$, $\tau_k(0) = 1$, $u^\star(-1)=\emptyset$.
\FOR {$n=0$ \TO $T$}
\STATE For all $k$, Update $\hat c_{k}(n)$, $\hat\theta_k(n)$, $\tau_k(n)$ and $t_{k}(n)$.
\STATE $u^\star(n) = u^\star(n-1) $;
\STATE \textbf{\underline{1. Updating the Leader             }}
\IF { $n$ mod $4 = 0$}

\FOR {$l=1$ \TO $L$} 
  \STATE $a_l(n) = \arg\max\limits_{k\notin \{a_i(n) : i<l\}} \hat c_k(n)$;
\ENDFOR 
\FOR {$l=1$ \TO $L$}
  \STATE $u^\star_l(n) = \arg\max\limits_{k\notin \{u^\star_i(n) : i<l\}, h(k)=h(a_l(n))} \hat\theta_k(n)$.
\ENDFOR
\ENDIF

\STATE $u(n) = u^\star(n)$
\STATE $W = n$ mod $4$
\STATE \textbf{\underline{2. Exploitating the Leader.          }}
\IF{ $W = 0$ or $3$}
    \STATE \textbf{if} $W=3$ \textbf{then} $u(n) = $shuffle$(u(n))$ \textbf{end if} 
    \STATE Play $u(n)$ and continue to round $n+1$.
\ENDIF
\STATE \textbf{\underline{3. Type-2 Exploration           }}
\IF{ $W = 1$}
\IF{$\exists k\notin u(n)$ and $l\leq K$: $b_k(n)>\hat\theta_{u_l(n)}(n)$ and $h(k) = h(u_l(n))$ } 
  \STATE $u_{l+1}(n) = u_l(n)$ $\forall l = 1\dots L-1$, $u_1(n) = k$. 
  \STATE Play $u(n)$ and continue to round $n+1$.
\ENDIF
\ENDIF
\STATE \textbf{\underline{4. Type-1 Exploration           }}
\IF{$\exists k:$ $d_k(n) > \hat c_{u_L(n)}(n)$ and $h(k)\neq h(u_L(n))$} 
  \STATE $u_L(n) = k$. 
    \STATE Play $u(n)$ and continue to round $n+1$.
  \ENDIF
\STATE Play $u(n)$.
\ENDFOR
\end{algorithmic}
\end{algorithm}

\subsection{Regret under LDR}

The next theorem provides a (finite-time) upper bound of the average number of times a suboptimal item $k$ ($k\notin u^\star$) is displayed, and from there, an asymptotic regret upper bound for LDR. To state these results, we define the list $y^k$ as the list obtained by replacing the last item in $u^\star$ by $k$, i.e., $\forall i<L$, $y^k_i=u_i^\star$ and $y_L^k=k$ and $w^k$ the list obtained by appending $k$ to the top of $u^\star$ (i.e. $w^k_1=k$ and $w^k_i = u^\star_{i-1}$ for all $i>1$). Recall that $u(n)$ denotes the list selected under LDR in round $n$.

\begin{theorem}\label{th:DiversiPIERegret}
There exists a constant $C(N,L,\delta)$ depending on $N, L$ and $\delta$, and a round $n_0$ such that under the LDR algorithm, we have: $\forall T>n_0$, $\delta>0$ and $\forall k>L$,
\begin{align*}
\EE[\sum_{n=1}^T\mathbbm{1}[k\in u(n)]\leq &\frac{f(T) }{ I(\nu(L|y^k) + \delta, \nu(L|u^\star))}+  \\ &\frac{f(T)}{I(\phi_{h(k)} \theta_{k h(k)}+\delta, \phi_{h(k)} \theta_{L_k h(k)}-\delta)}+\\ &C(N,L,\delta).
\end{align*}
where $L_k = \max\{l \leq L: h(u^\star_l) = h(k)\}$ and
\begin{align*}
C(N,L,\delta) \leq &[4N(N+2)\delta^{-1}+1][8L(2L+\delta^{-2})+\\&8L(N-L)(2(N-L)+\delta^{-2})] \\ & +(8L+24)(2+\delta^{-2})+\\ &8(N-L)(2(N-L)+\delta^{-2})+8L^2 (2L+\delta^{-2}) \\ &+ (4N(N+2)+1)C_1+ C_2.
\end{align*} 

Moreover:
\begin{align*}
\lim\sup\limits_{T\to\infty} \frac{R^{\text{LDR}}(T)}{\log
(T)} \leq& \sum\limits_{k>L} \frac{\mu(u^\star) - \mu(y^k)}{ I(\nu(L|y^k) + \delta, \nu(L|u^\star)-\delta)} +\\& \frac{\mu(u^\star) - \mu(w^k)}{I(\phi_{h(k)} \theta_{k h(k)}+\delta, \phi_{h(k)} \theta_{L_k h(k)}-\delta)}.
\end{align*}
\end{theorem}

We outline the main steps of proof below.

\medskip
\noindent
{\it Sketch of proof.} To derive an upper bound of $\EE[\sum_{n=1}^T \mathbbm{1}[k\in u(n)]]$, $\forall k>L$, we combine concentration-of-measure arguments and an appropriate decomposition of the set of rounds. We introduce the following sets of rounds:
\begin{align*}
B &= \{n: \exists k:\ b_k(n) < \phi_{h(k)} \theta_{k h(k)}\},\\
A^k & = \{n\notin B: k \in u^\star(n)\},\\
C^k & =\{n\notin \cup_{j>L} A^j\cup B : k\in u(n)\text{ and }d_k(n)>\hat{c}_L(n)\},\\
D^k & =\{n\notin \cup_{j>L} A^j\cup B : k\in u(n)\text{ and }\\ &\ \ \ \ \ b_k(n)\geq \hat\theta_{u^\star_l(n)}(n)\text{ for some }l:\ h(u^\star_l(n)) =h(k)\},
\end{align*}
so that we can show that $\sum_{n=1}^T \mathbbm{1}[k\in u(n)]]\le |B|+|\cup_{j>L}A^j|+|C^k|+|D^k|$. In fact, $C^k$ (resp. $D^k$) includes the set of rounds where $k$ in displayed using type-1 (resp. type-2) exploration. Now $B$ is the set of rounds where the index of a given item underestimates its true click-through-rate, and applying the concentration inequality of Theorem 1 in \cite{garivier2013informational}, $B$ is finite in expectation. The main difficulty of the proof is showing $u^\star$ is identified in finite time - i.e. $\EE[|A^k|]<\infty$. To this end we further split the rounds in $A^k$ as follows. For $\delta$ {small enough}, we first prove that $A^k\subset A_1^k\cup A_2^k$ where:
\begin{align*}
A^k_1& = \{n\in A^k: \exists y\in u^\star:h(y) = h(k)\text{ and }\\ &|\hat\theta_{y} - \phi_{h(k)}\theta_{yh(k)}|\geq \delta\text{ or }|\hat\theta_k(n) - \phi_{h(k)}\theta_{kh(k)}|\geq\delta\},
\end{align*}
\begin{align*}
A^k_2=\{n\in A^k\setminus A^k_1: \forall y \in u^\star \text{ with }&h(y) = h(k)\\&\text{ then }y \in u^\star(n)\}.
\end{align*}
We then use a concentration inequality (Lemma 5 in \cite{Combes2015LearningToRank}) to obtain $\EE[|A^k_1|]<\infty$ and hence LDR is able to correctly order items of the same topic in all but a finite number of rounds. Thus, $\forall k>L$ and $n\in A^k_2$, the expected \emph{success rate} for round $n$ satisfies $\EE[\hat c_k(n+1)t_k(n+1) - \hat c_k(n) t_k(n)]\leq \nu(L|y^k)$, i.e. the observations of the \emph{success rate} of $k$ at rounds $n\notin A^k_1$ are drawn from a distribution of mean at most $\nu(L|y^k) < \nu(L|u^\star)$. Consequently, $\hat c_k(n)$ will converge towards a value less than $\nu(L|y^k)<\nu(L|u^\star)$ and hence will eventually not be chosen in the leader $u^\star(n)$, due to the first step of LDR's procedure of computing the leader. Using a similar concentration argument as for bounding $\EE[|A^k_1|]$, we obtain $\EE[|A^k_2|]<\infty$. Since in all rounds outside $\cup_{k>L}A^k\cup B$ we have $u^\star = u^\star(n)$, we also have $\hat c_k(n)$ converges to an unbiased estimate of $\nu(L|y^k)$ as all items are explored infinitely many times and $\EE[|A^k\cup B|]<\infty$. The cardinalities of $C^k$ and $D^k$ are then easily bounded in light of the definition of the indexes $b_k(n)$ and $d_k(n)$, and using similar arguments as those used in the analysis of KL-UCB for the classical MAB problem \cite{garivier2011kl}.

\section{Regret Lower Bound}\label{sec:low}

In this section, we derive regret lower bounds satisfied by any list selection policy. To this aim, we can directly apply the generic methodology developed in \cite{graves1997asymptotically} in the broad context of controlled Markov chains. However the obtained regret lower bound would be the solution of an involved optimization problem; it would not be explicit, and even the way it scales as a function of $N$ and $L$ would be hard to guess. To circumvent this difficulty, we consider a slightly modified model, where the decision maker is getting some help from an Oracle. More precisely, the Oracle first reveals the distribution $\phi$. Then as in the original model, when a query arrives, its topic is unknown, and the decision maker returns a list. But after the list is displayed, the Oracle reveals the actual topic of the query. In turn, the decision maker has now additional information, since in the original model, the topic of the query is revealed only when the user actually clicks on one of the items of the list\footnote{It is tempting to propose a model where the Oracle reveals $\phi$ only, but this would again lead to an intractable regret lower bound.}. 

We derive a regret lower bound for the new model with the Oracle help, and show that it scales as $O(N\log(T))$. Of course, this bound also constitutes a regret lower bound for our original model, since in the new model, we can devise list selection policies that just ignore the additional information provided by the Oracle. Note that later on, we will devise a sequential list selection policy for our original model whose regret is provably upper bounded by $O(N\log(T))$. This implies that our regret lower bound is order-optimal. 

Let $\Theta=\{ \theta\in [0,1]^{N\times M}: \forall k,m, \theta_{km}=0 \text{ if }h(k)\neq m\}$ and let $\Lambda = \{\phi\in [0,1]^M: \sum_{m=1}^M\phi_m=1\}$. For a given $\phi\in \Lambda$, define the set of {\it bad} parameters $B(\theta) = \cup_{k>L} B_k(\theta)$ where
$
B_k(\theta) = \{\lambda\in\Theta : \exists i\leq L : u(\lambda, \phi)_i^* = k \text{ and }I^{u^\star}(\theta, \lambda) = 0\},
$
and where for any list $u$, $I^u(\theta,\lambda)$ is the Kullback-Leibler information between the  parameters $\theta$ and $\lambda$ as observed when the list $u$ is displayed (refer to \cite{graves1997asymptotically} for a detailed explanation):
$$
I^{u}(\theta, \lambda) = \sum\limits_{k=1}^{L} \phi_{h(k)} I(\theta_{u_kh(k)}, \lambda_{u_k h(k)}) \prod\limits_{s=1}^{k-1} (1 - \theta_{s h(k)}),
$$
where $I(a,b)=a\log(a/b)+(1-a)\log((1-a)/(1-b))$ is the KL divergence between two Bernoulli distributions of respective means $a$ and $b$. Recall that by convention,  the item $k>L$ is not listed in $u^\star$. Hence $B_k(\theta)$ can be interpreted as the set of parameters in $\Theta$ that can not be distinguished from $\theta$ when selecting $u^\star$, and such that $k$ is in the optimal list.

For all suboptimal items $k>L$, we define the list $u^k$ such that $u^k =\argmax_{u: k = u_1} \mu(u)$, the list containing item $k$ in the first slot with the highest expected reward. It is easy to see $u^k$ can be obtained through the procedure described in Section \ref{sec:optimalPolicy} (starting from $u^{[1]}=k$) and differs from $u^\star$ only in one element (either the very last element or the last element with the same topic as $k$ are removed from $u^\star$ and $k$ is added in the first slot). Finally we say that a policy $\pi\in \Pi$ is {\it uniformly good}, if for any parameters $(\phi,\theta)\in\Lambda\times \Theta$, its regret satisfies $R^\pi(T)=o(T^a)$ for all $a>0$ (we will show that our proposed policy LDR is uniformly good). We are now ready to state our regret lower bound.
\begin{theorem}\label{th:LowBoundRew1}
For any parameter $(\phi,\theta)\in\Lambda\times \Theta$, and for any uniformly good policy $\pi\in \Pi$, we have:
$$
\lim\inf_{T\to\infty} \frac{R^\pi(T)}{\log(T)} \ge \sum_{k > L} \frac{\mu(u^\star) - \mu(u^k)}{\min_{\lambda\in B_k(\theta)} I(\theta_{kh(k)},\lambda_{kh(k)})}.
$$
\end{theorem}
The above theorem is proved in Section \textsection\ref{subset:1}. It is a consequence of the more precise following statement indicating the minimum amount of times suboptimal lists should be explored. We establish that if the Oracle reveals $\phi$ and the topic of the queries at the end of each round, then under the best possible policy, the list $u^k$ should be displayed at least $\log(T)/\min_{\lambda\in B_k(\theta)} I(\theta_{kh(k)},\lambda_{kh(k)})$ times asymptotically, whereas any other sub-optimal list should be displayed at most $o(\log(T))$ times. Hence, with the help of the Oracle, we would just need to explore in the first slot of the displayed list. We believe that this statement would not hold in the original model, since $\phi$ needs to be learnt. That is why LDR, our sequential list selection algorithm, performs two types of exploration (one for ranking all items, and one for ranking items within the same topic). Finally note that the regret lower bound scales as $\Theta((N-L)\log(T))$.

\subsection{Proof of Theorem \ref{th:LowBoundRew1}}\label{subset:1}

As stated above we assume that $\phi$ is revealed, and that the topic of the query is revealed at the end of each round. To prove the theorem, we then use the formalism of controlled Markov chains developed in \cite{graves1997asymptotically}. More precisely, the states are represented by the topic $m$ of interest of the past query and the position $x$ of the first relevant article in the presented list. The possible control laws correspond to lists of items (here the control does not depend on the state as in \cite{graves1997asymptotically}). The transition probabilities between two states $(x,m_1)$ and $(y, m_2)$ under action $u$ and parameter $\theta$ are given by:
$$
p((x,m_1),(y, m_2)|\theta, u) =  \phi_{m_2} \theta_{u_ym_2}\prod\limits_{l=1}^{y-1} (1-\theta_{u_lm_2})
$$
Following \cite{graves1997asymptotically}, the Kullback-Leibler information number between two parameters $\theta$ and $\lambda$ given that the list $u$ is presented is:
$$
I^{u}(\theta, \lambda) = \sum\limits_{k=1}^{L} \phi_{h(k)} I(\theta_{u_kh(k)}, \lambda_{u_k h(k)}) \prod\limits_{s=1}^{k-1} (1 - \theta_{s h(k)}),
$$
Let $\Theta=\{\theta \in [0,1]^{N\times M}: \theta_{im}>0\text{ iff }h(i) = m\}$ and define the set of {\it bad} parameters $B(\theta) = \cup_{k>L} B_k(\theta)$ where
$$
B_k(\theta) = \{\lambda\in\Theta : \exists i\leq L : u(\lambda, \phi)_i^* = k \text{ and }I^{u^\star}(\theta, \lambda) = 0\},
$$
Note that we do not include $\phi$ as an unknown parameter since we assume that the Oracle is revealing $\phi$. Then, a direct application of the results in \cite{graves1997asymptotically} yields the following:

\medskip
\noindent
For any uniformly good algorithm $\pi$ we have:
$$
\lim\inf\limits_{T\to\infty} \frac{R^\pi(T)}{\log(T)} \geq c(\theta)
$$
where $c(\theta)$ is the minimal value of the following optimization problem:
\begin{equation}\label{eq:op}
\min_{c_u\ge 0, u\in \mathcal{U}} \sum\limits_{u\in\mathcal{U}} c_u(\mu(u^\star) - \mu(u))
\end{equation}
subject to:
\begin{equation}\label{eq:cc}
\min\limits_{\lambda \in B_k(\theta)} \sum\limits_{u\in\mathcal{U}} c_u I^u (\theta, \lambda)\geq 1,\ \forall k>L.
\end{equation}

Starting from this result, we can now proceed with the main argument of the proof. We assume the solution to the above optimization problem (\ref{eq:op})-(\ref{eq:cc}) is given by some $(c_g^*,g\in\mathcal{U})$. We show that if there exists $v\neq u^k$ for some $k>L$ such that $c_v^*>0$, we can create another solution $c$ where $c_v = 0$ and $c_{v'} = c^*_{v'}$ for all $v'\neq u^k$ that yields lower regret.

Let $c^*=(c_g^*, \ g\in\mathcal{U})$ be the solution of our optimization problem and assume there exists $v\neq u^i$ for some $i>L$ such that $c^*_v>0$. Define $c = (c_g,\ g\in\mathcal{U})$ as follows:
$$
c_g=
\begin{cases}
0,\text{ if }g=v \\
c^*_g + c^*_v\prod_{i<l} (1-\theta_{v_i h(k)}),\text{ if }\exists k:  g = u^k \text{ with }v_l =k\\
c^*_g, \text{ otherwise.}
\end{cases}
$$

We observe that $c$ satisfies the constraints (\ref{eq:cc}). Indeed:
$$
c_v I^v(\theta, \lambda) = c_v\sum\limits_{l=1}^L \phi_{h(v_l)} I(\theta_{v_l,h(v_l)}, \lambda_{v_l,h(v_l)}) \prod\limits_{s=1}^{l-1} (1-\theta_{v_s, h(v_l)})
$$
and by definition $\phi_{h(v_l)}I(\theta_{v_l,h(v_l)}, \lambda_{v_l,h(v_l)}) = I^{u^{v_l}}(\theta, \lambda)$. We can then rewrite the above term as:
$$
c_v I^v(\theta, \lambda) = \sum\limits_{l=1}^L I^{u^{v_l}}(\theta, \lambda) c_v \prod\limits_{s=1}^{l-1} (1-\theta_{v_s, h(v_l)})
$$
Thus the values of the constraints are identical under $c^*$ and $c$.

Let us denote by $c^*(\theta)$ and $c(\theta)$ the values of the objective function (\ref{eq:op}) under $c^*$ and $c$, respectively.  We show that $c^*(\theta) - c(\theta) \geq 0$. By our definition of $c^*$ and $c$ we have:
$$
c^*(\theta) - c(\theta) = c_v(\mu^\star - \mu(v)) - c_v\sum\limits_{i=1}^L \prod\limits_{s=1}^{i-1} (1-\theta_{v_s, h(v_i)}) (\mu^\star - \mu(u^{v_i})).
$$
We then have $c^*(\theta) \geq c(\theta)$ if:
\begin{align*}
&\sum\limits_{m=1}^M   \phi_m\sum\limits_{l=1}^L\left[\prod\limits_{s=1}^{l}(1-\theta_{v_sm}) - \prod\limits_{s=1}^{l}(1-\theta_{u^\star_sm})\right] \geq \\ &\sum\limits_{l=1}^{L} \prod\limits_{s=1}^{l-1}(1-\theta_{v_s h(v_l)}) \sum\limits_{m=1}^M\phi_m\left[\prod\limits_{s=1}^{L}(1-\theta_{u^{v_l}_sm}) - \prod\limits_{s=1}^{L}(1-\theta_{u^\star_sm}) \right]
\end{align*}
Now the above inequality will be a consequence of: for all $l$,
\begin{align*}
&\sum\limits_{m=1}^M   \phi_m\left[\prod\limits_{s=1}^{l}(1-\theta_{v_sm}) - \prod\limits_{s=1}^{l}(1-\theta_{u^\star_sm})\right] \geq \\ &\prod\limits_{s=1}^{l-1}(1-\theta_{v_s h(v_l)}) \sum\limits_{m=1}^M\phi_m\left[\prod\limits_{s=1}^{L}(1-\theta_{u^{v_l}_sm}) - \prod\limits_{s=1}^{L}(1-\theta_{u^\star_sm}) \right].
\end{align*}
We establish the latter inequality, by observing that:
\begin{align*}
&\sum\limits_{m=1}^M  \phi_m\left[\prod\limits_{s=1}^{l}(1-\theta_{v_sm}) - \prod\limits_{s=1}^{l}(1-\theta_{u^\star_sm})\right] \geq
\\ &\prod\limits_{s=1}^{l-1}(1-\theta_{v_s h(v_l)})\sum\limits_{m=1}^M \phi_m\left[\prod\limits_{s=1}^{l}(1-\theta_{u^{v_l}_sm}) -\prod\limits_{s=1}^{l}(1-\theta_{u^\star_sm})\right] \geq
\\ &\prod\limits_{s=1}^{l-1}(1-\theta_{v_s h(v_l)})\sum\limits_{m=1}^M\phi_m\left[\prod\limits_{s=1}^{L}(1-\theta_{u^{v_l}_sm}) - \prod\limits_{s=1}^{L}(1-\theta_{u^\star_sm}) \right].
\end{align*}
The last inequality is due tot the fact that all permutations of a list $u$ have the same expected reward, without loss of generality, for each slot we consider $u^\star$ such that $u^\star_i = u_i^{v_l}$ for all $i>1$. The first inequality is due to $\prod\limits_{s=1}^{l-1}(1-\theta_{v_s h(v_l)})<1$ and
$$
\sum\limits_{m=1}^M \phi_m\prod\limits_{s=1}^{l}(1-\theta_{u^{v_l}_sm}) \leq \sum\limits_{m=1}^M \phi_m\prod\limits_{s=1}^{l}(1-\theta_{v_s m})
$$
which is a consequence of our construction of $u^{v_l}$ according to the procedure in Section \ref{sec:optimalPolicy} - i.e. the probability of a user not clicking on an item above slot $l$ among policies containing $v_l$ is minimized by $u^{v_l}$. \ep

\section{Numerical Experiments}\label{sec:num}

To evaluate the performance of LDR and compare with that of other algorithms, we use both randomly generated data and a model constructed from user click data obtained from the music search engine of Spotify. {In this section, we define the regret of a run of an algorithm as the reward accrued by the optimal policy minus the reward obtained by the algorithm.}

\subsection{Datasets}

\noindent
{\bf Artificial Data.} We first measure the performance of our algorithms on artificially generated models. To this end, we generate $20$ random problem instances using the parameters $L=10$ result slots, $M=5$ topics and $N=40$ items. For each problem instance, the CTRs $\theta_{kh(k)}$ for $k\in \mathcal{ N}$ are chosen uniformly at random in $[.2,1]$, and we simulate the algorithms using several runs so as to get 90\% confidence interval on their regrets.

\medskip
\noindent
{\bf Real Data.} To build a realistic statistical model of user behavior from click-through data, we use the so-called \emph{Cascade Model} proposed in \cite{craswell2008experimental}. Our dataset contains $74631$ data points, consisting of click information of users searching the album repository (we select the most popular 85 albums, which are classified into 10 genres), issuing the query "Christmas" in the first half of December 2015. For each query, we have collected the item that was clicked, its position in the list and its genre, obtained from meta-data. Since during the period of time the data was collected, the list of albums displayed to the users were not heavily personalized, we obtain an \emph{average} list that was displayed to users  by ordering the albums in ascending order of their average position where they were selected. We then construct the cascading model by splitting the entries by topic and computing $\theta_{kh(k)}$ as the number of times $k$ was clicked divided by the sum between the total number of clicks occurring on items of the same topic placed below (and including) $k$ and the abandonment. We estimate the abandonment for our given query at $20\%$, however different values may be obtained depending on measures and use cases (time-out period, departure of a user from the system, activation of other service features, etc). Note that, in our experiments, variations in the choice of abandonment do not significantly change the outcome of our simulations in terms of the relative regret of the algorithms.

\subsection{Algorithms}

We use several baselines to gauge the efficiency of our algorithm: RBA (Ranked KL-UCB, presented in \cite{kohli2013fast}), PIE (presented in \cite{Combes2015LearningToRank}) and, in the case of the real data experiments, a policy that always selects the most popular albums (as measured by Spotify). We chose Ranked KL-UCB, as it is the current state of the art in the settings that account for diversity, PIE for its simplicity and efficiency in its original setting and popularity based ranking as this is most often used in practice.

\medskip
\noindent
{\it LDR.} For ease of implementation and increased numerical performance we slightly modify the algorithm presented in Section \ref{sec:algo}. In our implementation of LDR, $W$ is an integer chosen in $[0,3]$ uniformly at random in each round (instead of $W = n\mod 4$) and we update the leader $u^\star(n)$ at every round.

\medskip
\noindent
{\it PIE*.} PIE* is a slight variation of the PIE(L) algorithm presented in \cite{Combes2015LearningToRank} that shows better numerical performance in our considered setting. PIE* simply collects observations about an item only when it is inspected and builds a KL-UCB index for each item using these observations. It then selects the items with the $L$ highest KL-UCB indexes. As it turns out, PIE* then explores only in the last slot: the items selected a number of rounds growing logarithmically with time are only those shown in the last slot. Note that even though the reward of a click is the same regardless of the slot in which it occurs, PIE(L) outperforms PIE(1) due to the diversity requirements of our setting which are not present in \cite{Combes2015LearningToRank}. More importantly, since PIE* explores in the last slot only, it cannot rank items within each topic (as this is done under LDR using type-2 exploration) -- as consequence PIE* (or actually any algorithm presented in \cite{Combes2015LearningToRank}) is not uniformly good, i.e., there are problem instances where its regret grows linearly with time as we show later in this section.

\medskip
\noindent
{\it RBA.} The RBA algorithm is presented in \cite{kohli2013fast} and runs a bandit algorithm for each slot, each such algorithm only considering the plays occurring in its assigned slot. In our case, we chose to run an instance of KL-UCB per slot. This algorithm is proven to have an expected regret scaling as $O(LN\log(T))$.

\subsection{Experiment Results} 

\begin{figure}[ht!]%
    \centering
      \includegraphics[width=0.455\textwidth]{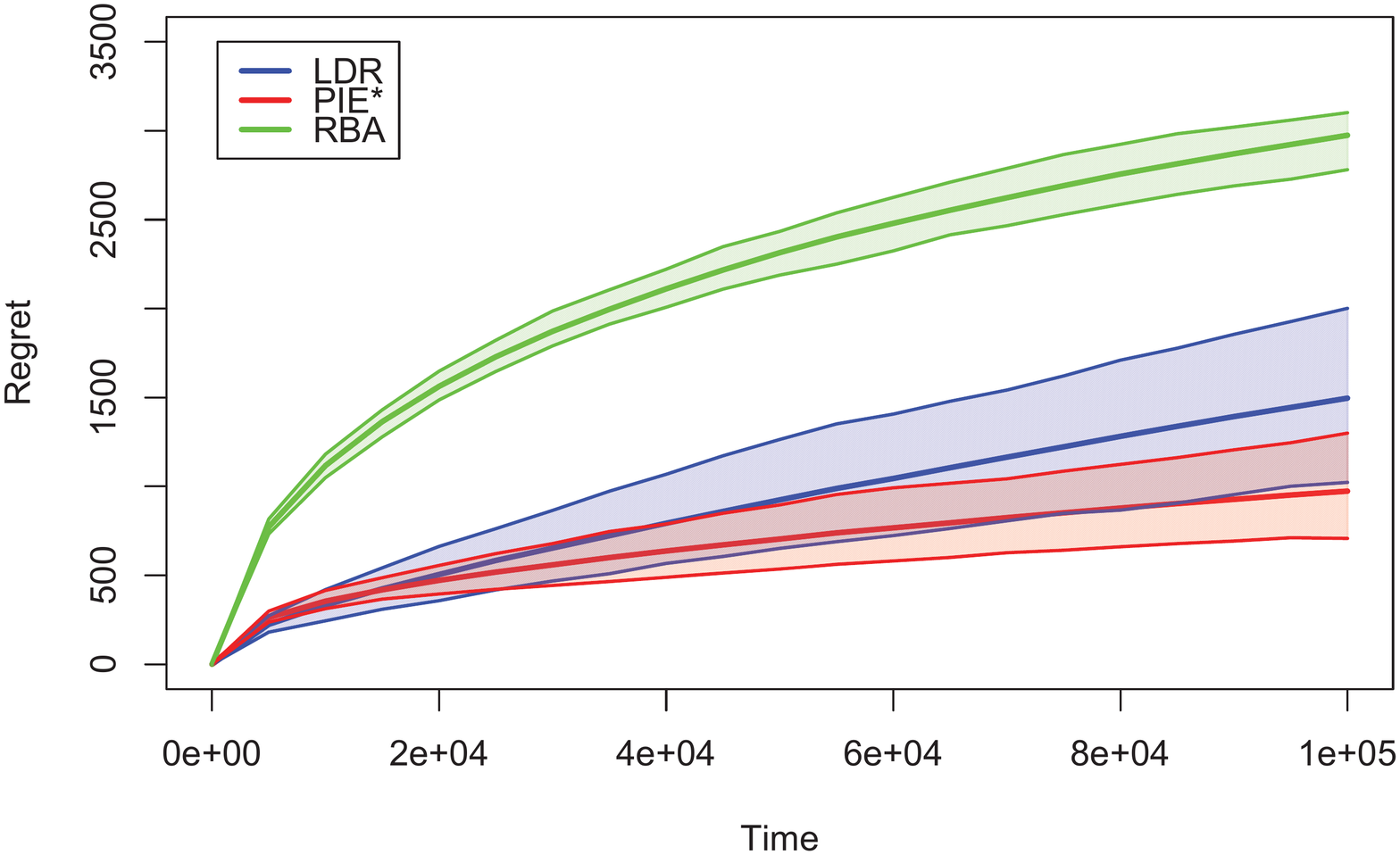} 
  \includegraphics[width=0.455\textwidth]{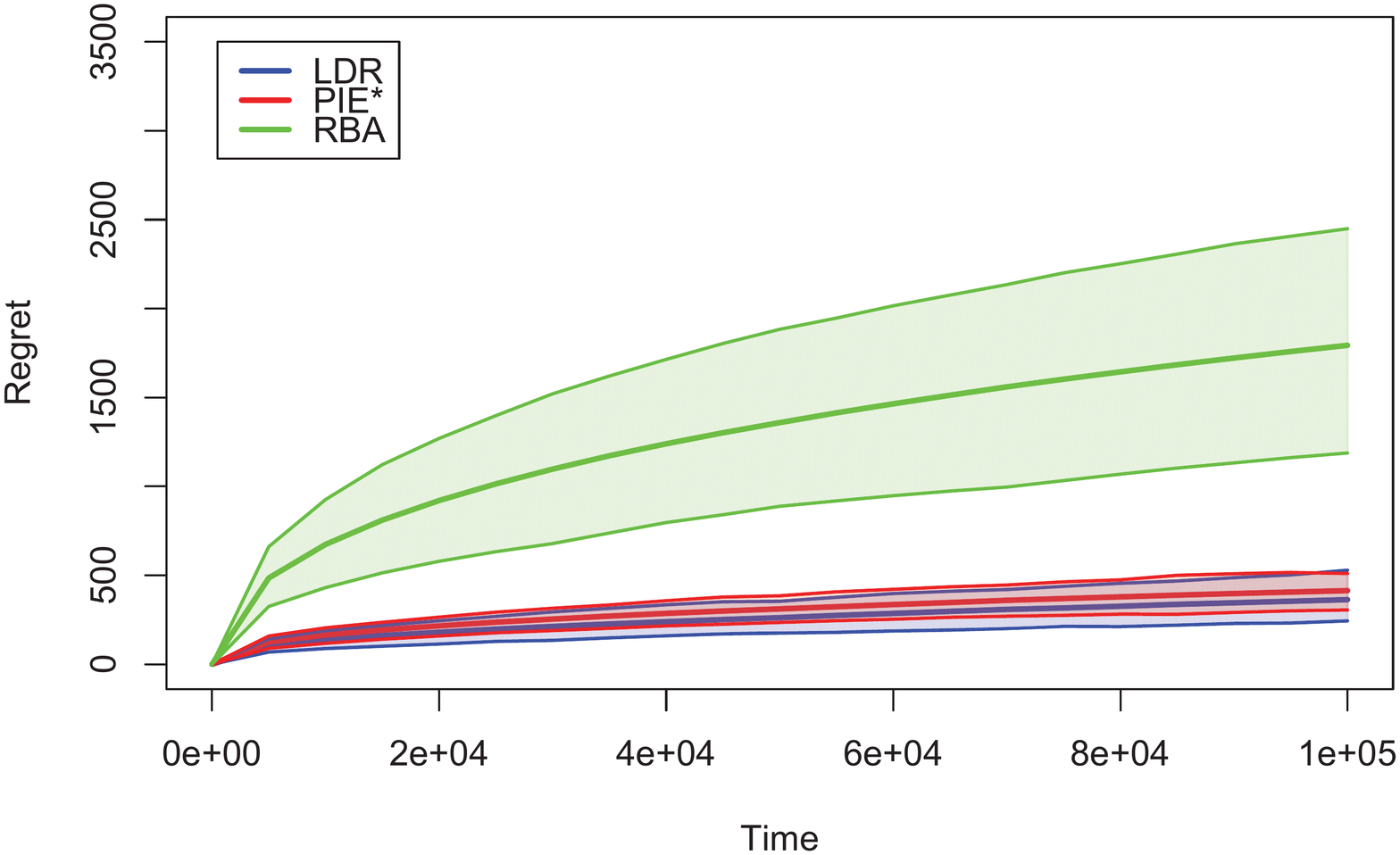} 
    \caption{Average regret and 90\% {quantiles} vs. time of the various algorithms on artificial data -- averaged over 20 randomly generated models. In the left figure $N=40$, $L=10$ and in the right figure $N=50$, $L=20$.}%
    \label{fig:example}%
\end{figure}

\begin{figure}[ht!]%
    \centering
  {\includegraphics[width=0.455\textwidth]{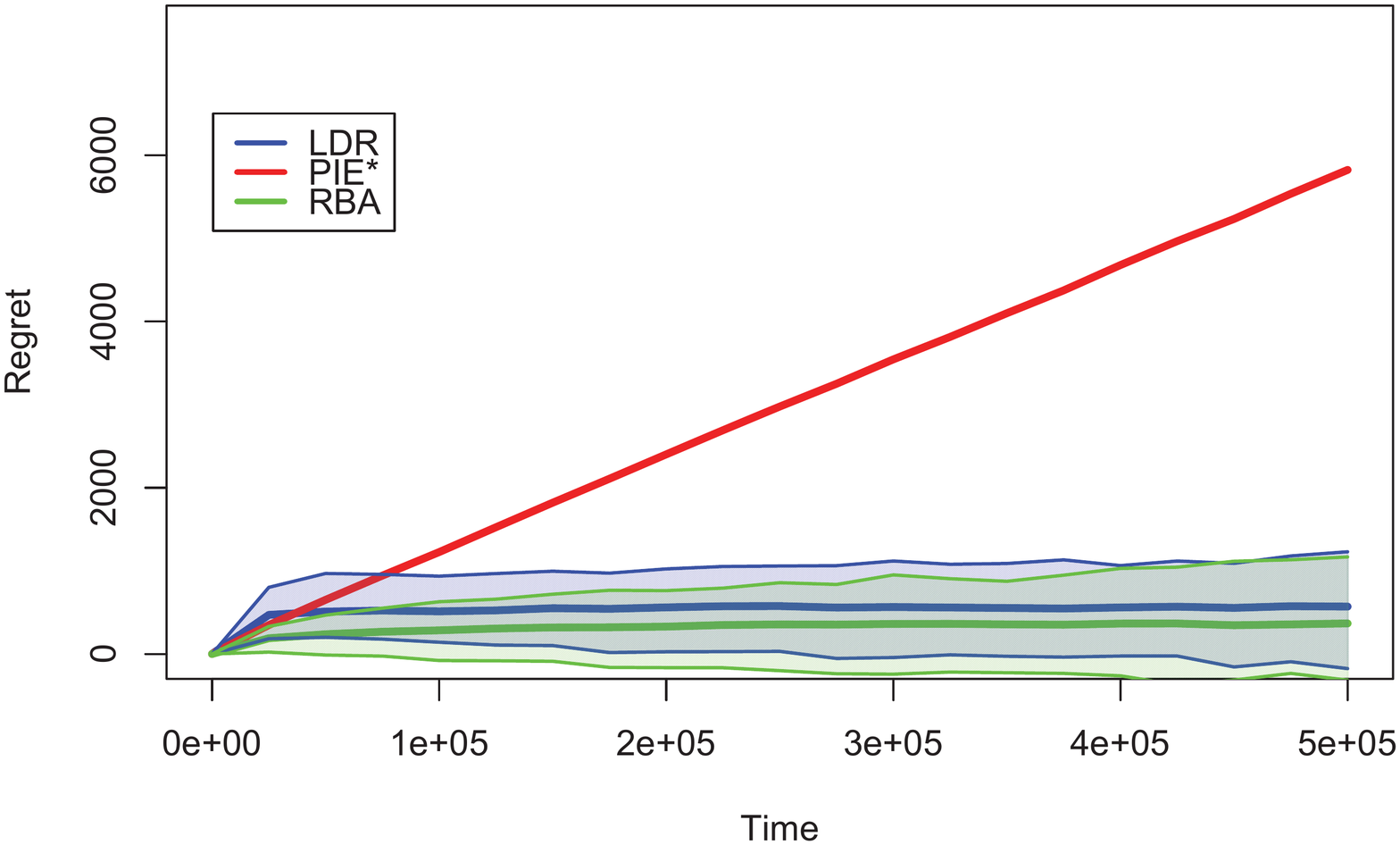} }%
     {\includegraphics[width=0.455\textwidth]{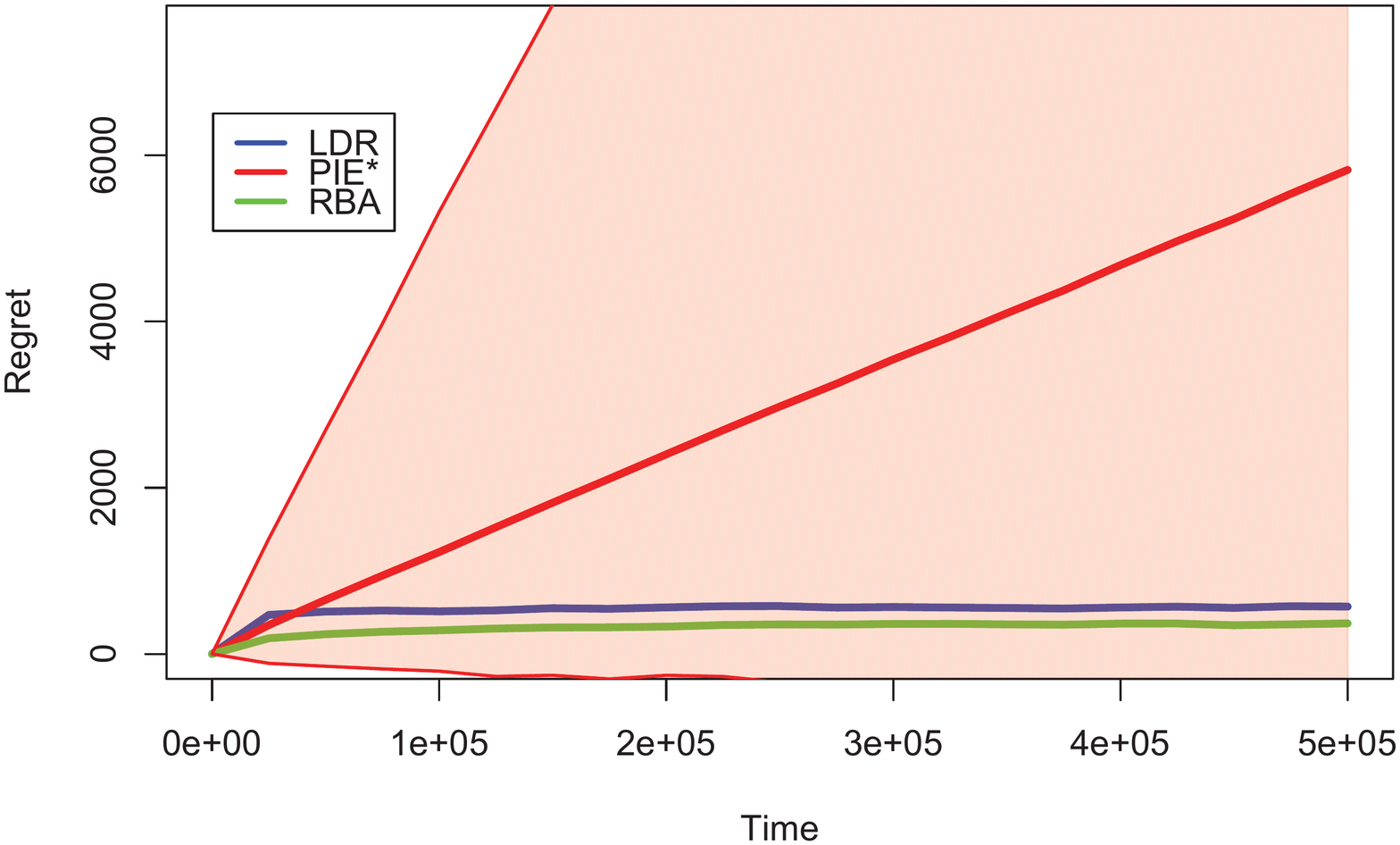} }%

    \caption{Regret vs. time of various algorithms -- a problem instance where PIE* yields linear regret. For clarity, we provide the confidence intervals for LDR and RBA in the left figure, and that of PIE* in the right figure.}
    \label{fig:example2}
\end{figure}

\noindent
{\bf Artificial data.} We present in Figure \ref{fig:example} the regret and its 90\% {quantiles} under the three algorithms for the problem instances described above. As expected RBA is outperformed by LDR. This performance gap is to be expected given the regret of RBA is suspected to scale as $O(LN\log(T))$ while we showed the regret of LDR scales as $O((N-L)\log(T))$. Surprisingly, {in the top plot,} PIE* outperforms all other algorithms, having a very narrow edge over LDR. Despite not being designed to account for diversity, PIE* does so nonetheless due to the intrinsic submodularity of \emph{success rates} of items belonging to the same topic. Using many items of the same topic will result in progressively lower \emph{success rates}. Hence items placed lower in the list will be identified as redundant and replaced by items from less used topics. In the bottom plot of Figure \ref{fig:example} we notice that PIE* no longer outperforms LDR (when $L=20$, $N=50$). It is important to also note that PIE* is not \emph{uniformly good} - i.e. it is not guaranteed to play the suboptimal rankings sublinearly in $T$ in all problem instances. PIE* can be viewed as a reckless version of LDR. LDR employs an additional exploration phase that guarantees it eventually identifies the optimal list w.p. 1 by correctly ordering items of the same topic. 

\medskip
\noindent
{\bf Regret of PIE* linearly growing in $T$.} Next we present a problem instance where PIE* exhibits linear regret. In Figure \ref{fig:example2}, we simulated PIE*, LDR and RBA on a toy example: we have $L=2$, two topics containing two items each, arriving with probability $1/2$ and $\theta_{11}=0.9, \theta_{21} = 0.8, \theta_{32} = 0.35$ and $\theta_{42}=0.3$. In this simple example, w.p. > 0, PIE* cannot correctly order the items in topic $1$ and appears to generate linear regret while both LDR and RBA display regret scaling as $O(\log(T))$. Note that the RBA and LDR have very similar regret as $L=2$ is small. Further observe that the regret under PIE* is highly variable. This is due to the fact that by chance, PIE* may initially guess the right ordering within each topic, in which case the regret is very low. PIE* exhibits very high regret when the ordering within topics is wrong.

\begin{figure}[ht!]%
    \centering
    \includegraphics[width=0.455\textwidth]{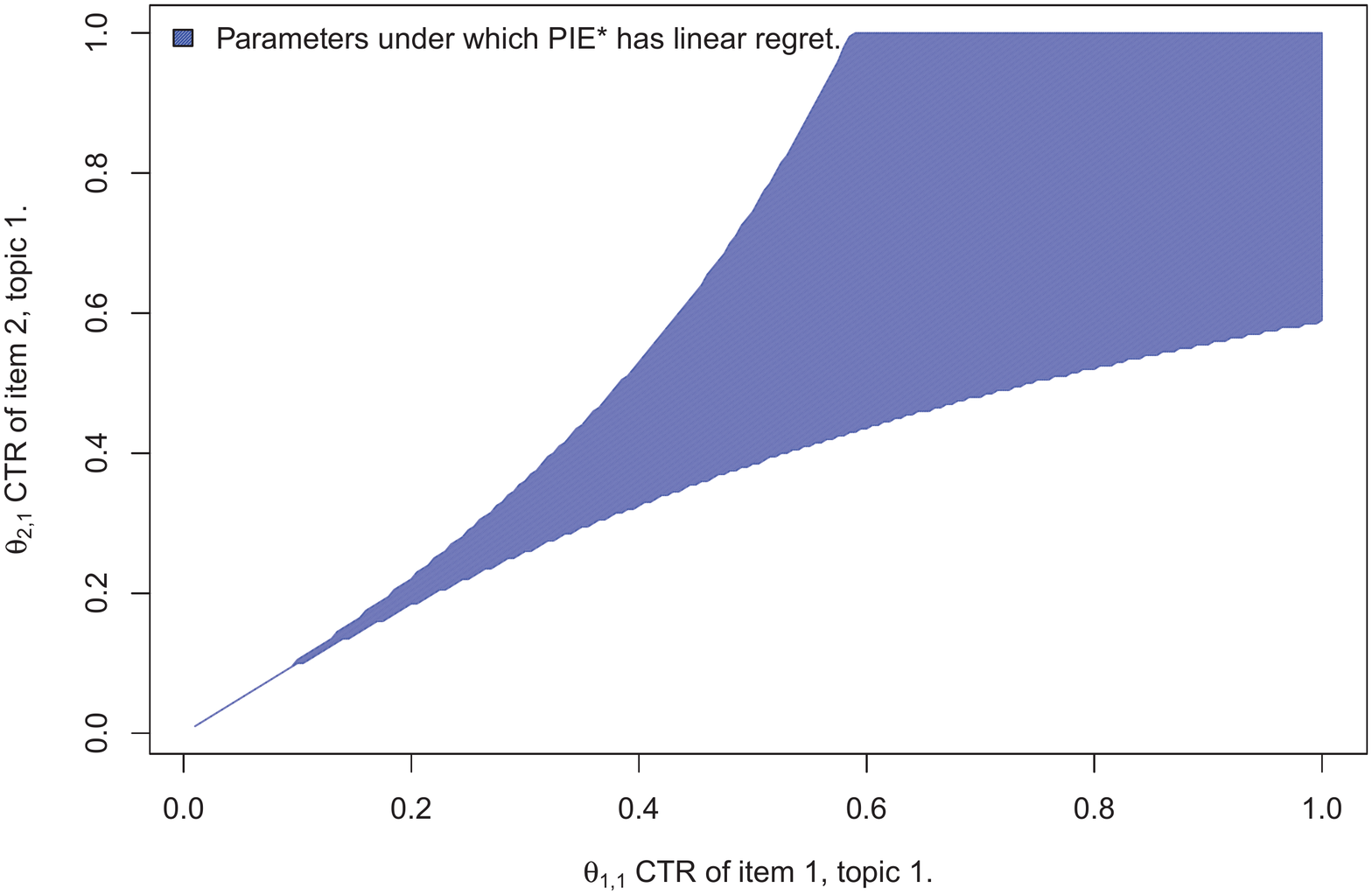} 
    \caption{CTR of items in topic $1$ for which PIE* has linear expected regret i.e. $ \phi_1 \theta_{2,1}\geq \phi_1(1-\theta_{2,1})\theta_{1,1}/(1-\phi_1\theta_{2,1})$}%
    \label{fig:params}%
\end{figure}
\begin{figure}[ht!]%
    \centering
   \includegraphics[width=0.455\textwidth]{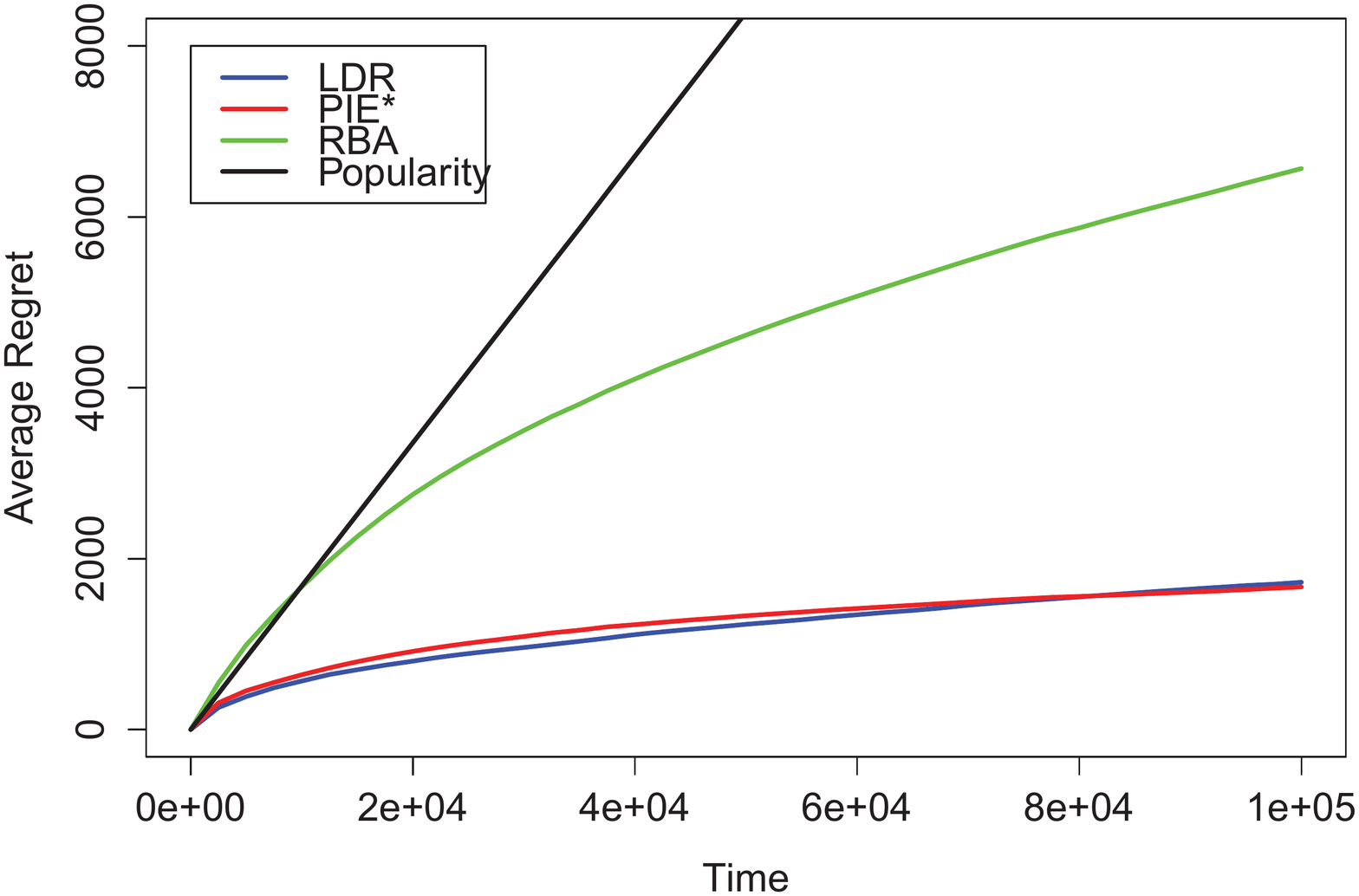}
    \caption{Regret (averaged over 100 runs) vs. time of the various algorithms on real-world data.}%
    \label{fig:example3}%
\end{figure}

 {It is important to note that PIE* is \emph{not} uniformly efficient, and while it might appear competitive relative to LDR in some problem instances, this is not always the case. Furthermore, it is impossible to know whether PIE* will have linear regret or not without knowing the problem parameters. In Figure \ref{fig:params}, the colored region represents the parameters under which two items of topic $m$ are presented in the incorrect order by PIE* with positive probability (in this plot, we consider $\phi_1 = 0.5$). As we can see, a significant fraction of the parameter space satisfies this condition, particularly, when the click-through-rates of the two items are similar. More precisely, if two items $i$ and $j$ (assume $\theta_{im}>\theta_{jm}$) belonging to a topic $m$ satisfy:
$$
\phi_m \theta_{jm} > \frac{\phi_m(1-\theta_{jm})\theta_{im} }{1-\phi_m\theta_{jm}},
$$ 
then, PIE* will not correctly order the two items with positive probability, and hence, its expected regret will scale linearly with time. Intuitively, the left hand side of the inequality represents the ratio of clicks to observations of item $j$ when in the first position, while the right hand side represents the ratio of clicks to observations of item $i$ when placed behind item $j$, in the second position.

Compared to PIE*, LDR is uniformly good, i.e., it performs well for all parameter distributions. Further remember that the regret of LDR scales as $O((N-L)\log(T)$, whereas the regret of RBA scales as $O(NL\log(T))$. This scaling is obvious from Figures \ref{fig:example3} and \ref{fig:example}, where $L>2$.}

\medskip
\noindent
{\bf Real data.} We present in Figure \ref{fig:example3} the regret under the three algorithms on the model built on real data. We also present the regret obtained by just ranking items according to their overall popularity. As expected, ranking by popularity performs poorly relative to presenting an optimally diverse list. As for artificial data, LDR and PIE* exhibit similar regret (PIE* makes the right guess for ordering items within topics on this problem instance), and outperforms RBA.

\section{Conclusion}

In this paper, we investigated the design of online learning-to-rank algorithms for systems answering users' queries by listing a few items selected from many. The originality and practical relevance of our model lie in the fact that items are categorized into topics, and that the topic of an arriving query is not known a priori. As a consequence, the system should output a list adhering to the diversity principle, i.e., covering several topics. For this difficult online learning problem, we have derived fundamental performance limits (regret lower bounds) satisfied by any algorithm, and have proposed LDR, an algorithm matching order-wise these limits. In our model, the topic of the various items is known, which is the case in practice in music search engines, our motivating application; in other systems, this might not be the case, and it is interesting to study whether the clustering of items into topics can also be learnt in an online manner.

\clearpage
\appendix
\section{Proof of Theorem 3.1}

 In the proof, we use the following result presented in \cite{Combes2015LearningToRank}. First, let us introduce the following notations. For all $T\in\NN$, define the random variables $o(T)$ (Bernoulli, independent) and $X(T)$ (Bernoulli i.i.d., $\EE[X(T)] = \chi$). Define $\mathcal{F}_T = \sigma(o(n), X(n), n\leq T)$ and $t(T) = \sum_{n\leq T} \mathbbm{1}[o(n) = 1]$. Further, let $\hat\chi(T) =\sum_{t\leq n} \mathbbm{1}[X(t)o(t)] /t(T)$.
 
 \begin{lemma}\label{lem:deviation}
Let us fix $c > 0$. Consider a random set of rounds $H \subset \NN$, such that, for all $n$, $\indic\{n \in H\}$ is $\mathcal{F}_{n-1}$ measurable. Further assume for all $n$ we have: $\EE[ o(n) | n \in H ] \geq c > 0$. Consider a random set $\Lambda = \cup_{s \geq 1} \{ \tau_s \} \subset \NN$, where for all $s$, $\tau_s$ is a stopping time such that $\sum_{n=1}^{\tau_s} \mathbbm{1} \{ n \in H \} \geq s$. 

  Then for all $\epsilon > 0$ and $\delta > 0$ we have that:

\eqs{
\sum_{n \geq 0} \PP[ n \in \Lambda , | \hat\chi(n) - \chi | \geq \delta] \leq  2 c^{-1} \left[ 2 c^{-1} +  \delta^{-2}  \right].
}
\end{lemma}

The proof consists in deriving an upper bound of $\EE[\sum_{n=1}^T \mathbbm{1}[k\in u(n)]]$. To this aim, we introduce the following sets of rounds:
\begin{align*}
B &= \{n: \exists k:\ b_k(n) < \phi_{h(k)} \theta_{k h(k)}\},\\
A^k & = \{n\notin B: k \in u^\star(n)\},\\
C^k & =\{n\notin \cup_{j>L} A^j\cup B : k\in u(n)\text{ and }d_k(n)>\hat{c}_L(n)\},\\
D^k& =\{n\notin \cup_{j>L} A^j\cup B : k\in u(n)\text{ and }b_k(n)\geq \hat\theta_{u^\star_l(n)}(n)\\ &\text{ for some }l:\ h(u^\star_l(n)) =h(k)\}.
\end{align*}
Let $k>L$ be a suboptimal item (i.e., $k\notin u^\star$). We first establish that: $\sum_{n\leq T} \mathbbm{1}[k\in u(n)] \leq |B|+|\cup_{j>L} A^j| + |C^k|+ |D^k|$. Let $n$ such that $k\in u(n)$, and assume that $n\notin B$. We distinguish two cases $k\in u^\star(n)$ or $k\notin u^\star(n)$. In the former case, $n\in A^k$. In the latter case, $k\notin u^\star(n)$ means that the algorithm explores in round $n$. There are two kinds of exploration corresponding to Events 2 and 3, respectively, as described in the algorithm. For Event 2 to occur, we need that for some $l$, $b_k(n)\geq \hat\theta_{u^\star_l(n)}(n)$ and $h(u^\star_l(n)) =h(k)$, and hence, if $n\notin  \cup_{j>L} A^j\cup B$, then $n\in D^k$. For Event 3 to occur, if $n \notin  \cup_{j>L} A^j\cup B$, we need $d_k(n) > \hat{c}_L(n)$ (indeed, note that since $n \notin  \cup_{j>L} A^j\cup B$, no suboptimal item is in the leader, and hence $u^\star=u^\star(n)$, and $L=u_L^*(n)$). This implies that $n\in C^k$.

Next we provide upper bounds on the cardinalities of the aforementioned sets of rounds. \\

\noindent\underline{(i) Upper bound of $\EE[ |B|]$.} Observe that $B=\cup_k B_k$ where $B_k=\{ n: b_k(n) < \phi_{h(k)}\theta_{kh(k)}\}$. Now when $b_k(n)< \phi_{h(k)}\theta_{kh(k)}$, by definition, we must have $\tau_k(n) I(\hat\theta_k(n), \phi_{h(k)}\theta_{kh(k)}) \ge f(n)$. Hence $B_k \subset \{ n: \tau_k(n) I(\hat\theta_k(n), \phi_{h(k)}\theta_{kh(k)}) \ge f(n)\}$. We deduce that:
\begin{align*}
\EE[ |B_k | ] & \le \sum_{n\ge 1} \PP [\tau_k(n)I(\hat\theta_k(n), \phi_{h(k)}\theta_{kh(k)}) \ge f(n)]\\
& \le \sum_{n\ge 1} 2e \lceil f(n)\log(n)\rceil \exp(-f(n))\\
&\le C_1,
\end{align*}
where the second inequality is obtained by applying the concentration inequality derived in Theorem 1 in \cite{garivier2013informational}, and the last inequality is due to our choice of $f(n)$ (which yields a converging Bertrand series). We get $\EE[ |B | ]  \le NC_1$.

\medskip
\noindent 
\underline{(ii) Upper bound of $\EE[ |A^k |]$.} Let us define:
\begin{align*}
\Delta=& \sup \{z>0: |\sum\limits_{m=1}^M \phi_m\theta_{km}-\phi_m\theta_{k'm}|\geq 4z\\&\text{ and }|\nu(L|y^k) - \nu(L|y^{k'})|\geq 4z, \forall k',k\geq L\},
\end{align*}
and let us fix $\delta \in (0,\Delta)$. We first prove that $A^k\subset A_1^k\cup A_2^k$ where 
\begin{align*}
A^k_1 = &\{n\in A^k: \exists y\in u^\star:h(y) = h(k)\text{ and }\\&|\hat\theta_{y} - \phi_{h(k)}\theta_{yh(k)}|\geq \delta\text{ or }|\hat\theta_k(n) - \phi_{h(k)}\theta_{kh(k)}|\geq\delta\},
\end{align*}
\begin{align*}
A^k_2=\{&n\in A^k\setminus A^k_1: \nexists y \in u^\star\text{ such that }h(y) = h(k)\\&\text{ and }y \notin u^\star(n)\}.
\end{align*}
Let $n\in A^k$. Then $k\in u^\star(n)$. Assume that $n\notin A_1^k$. Then for all $y\in u^\star$ such that $h(y)=h(k)$, $\phi_{h(k)}\theta_{yh(k)}$ and $\phi_{h(k)}\theta_{kh(k)}$ are {\it well} estimated, i.e., $|\hat\theta_{y} - \phi_{h(k)}\theta_{yh(k)}|< \delta$ and $|\hat\theta_k(n) - \phi_{h(k)}\theta_{kh(k)}|< \delta$. Then by our choice of $\delta$ and in view of the construction of $u^\star(n)$, all items in $u^\star$ with topic $h(k)$ are in $u^\star(n)$ since $k\in u^\star(n)$. Hence $n\in A_2^k$.

 

Next we proceed by deriving upper bounds of the cardinalities of $A_1^k$ and $A_2^k$. \\
\underline{(ii).1. Upper bound of $\EE[ |A_1^k|]$.} We first note that $A_1^k\subset E_1\cup E_2$ where
\begin{align*}
E_1 &= \{n\in A^k : |\hat\theta_k(n) - \phi_{h(k)}\theta_{kh(k)}|\geq\delta \},\\
E_2 &= \cup_{y\in u^\star: h(y)=h(k)} F^y,
\end{align*}
and 
\begin{align*}
F^y = \{ n\in A^k: &|\hat\theta_{y} - \phi_{h(k)}\theta_{yh(k)}|\geq \delta, \\&|\hat\theta_k(n) - \phi_{h(k)}\theta_{kh(k)}|< \delta \},
\end{align*}

We apply Lemma \ref{lem:deviation} to bound $\EE[|E_1|]$. In this lemma, we choose $i=k$, $H=\Lambda = \{n: k\in u(n)\text{ and } k=u_1(n)\text{ or }(n\mod 4 = 3\text{ and }k\in u^\star(n))\}$ and $c=1/L$. The stopping time $\tau_s$ corresponds to the time after which there have been $s$ rounds in $H$. Denote by $H_1=H\cap\{n:|\hat\theta_k(n) - \phi_{h(k)}\theta_{kh(k)}|\geq \delta\}$. We now proceed to show that $|E_1|\leq 4|H_1|$. For every round $n$ define $r(n)=\{m\in \NN: m\in [n-(n\mod 4),n - (n\mod 4)+3]\}$, the set of rounds between two updates of $u^\star(n)$ containing $n$. To this end we show that for every round in $E_1$, there exists some round $n'\in r(n)$ such that $n' \in H_1$. For every round $n\in E_1$ (hence $k\in u^\star(n)$) we then have that $|\hat\theta_k(n) - \phi_{h(k)}\theta_{kh(k)}|\geq \delta$ and denoting by $m = \max r(n)$ the last round in $r(n)$ (corresponding to the shuffling phase) we distinguish the following two cases:

a) $m \in E_1$ and consequently $\max r(n) \in H_1$.

b) $m \notin E_1$ and hence $\exists n'\in r(n)$ ($n'>n$) such that $n'\in E_1$ (and $|\hat\theta_k(n') - \phi_{h(k)}\theta_{kh(k)}|\geq \delta$) and $u_1(n') = k$ ($\tau_k(n')$ is incremented) and hence $n'\in H_1$. If such an $n'$ did not exist, we would have $\hat\theta_k(m) = \hat\theta_k(n)$, as no plays of $k$ occur between $n$ and $m$, and hence $m\in E_1$ a contradiction.

Hence, we have $|E_1|\leq 4|H|$. We get: $\EE[|E_1|] \le 8L(2L+\delta^{-2})$. 

Next, we proceed to bound the expected cardinalities of $F^y$, for all $y\leq L$. Note that when for all $n\in A^k$, $b_y(n)>\hat\theta_k(n)$ and $y$ is therefore a candidate for type-2 exploration if $y\notin u^\star(n)$. In Lemma \ref{lem:deviation} we choose $i=y$, and $H$ the set of rounds $n$ where $\tau_y(n)$ has a strictly positive probability of being incremented:
\begin{align*}
H = &\Lambda =\{n: k\in u^\star(n),\ y = u^\star_1(n)\text{ or } \\&(y\in u^\star(n)\text{ and }n\mod 4=3)\text{ or }\\&(b_y(n)>\hat\theta_k(n)\text{ and }n \mod 4=1\text{ and }y\notin u^\star(n))\}
\end{align*}
and $c = 1/(N-L)$ (assuming $N-L > L$, otherwise we can just take the smaller of the two). Using the same argument as above, we obtain $\EE[|E_2|] = 8 (N-L)(2(N-L) + \delta^{-2})$ and hence: 

$$
\EE[ |A_1^k|]\le 8L(2L+\delta^{-2})+8L(N-L)(2(N-L)+\delta^{-2}).
$$

\noindent\underline{(ii).2. Upper bound of $\EE[ |A_2^k|]$.}

Note that when $n\in A^k\setminus A^k_1$ we have:
$$
\EE[t_k(n+1) \hat c_k(n+1) - t_k(n) \hat c_k(n)] \leq \nu(L| y^k)
$$

Now we bound the number of rounds when $\hat c_k(n) - \nu(L| y^k)>2\delta$. To this end, let us split the rounds in $A^k_2$ into the following two sets:
$$
M = \{n\in A^k_2: t_k(n)<  (|\cup_{j>L} A^j_1|+|B|)/\delta\}
$$
$$
K = \{n\in A^k_2: t_k(n) \geq (|\cup_{j>L} A^j_1|+|B|)/\delta\}.
$$
and
$$
K' = \{n\in K: \hat c_k(n) \leq \nu(L|y^k) + \delta\}.
$$

\noindent\underline{(ii).2.a) Expected cardinality of $M$}: If $n\in M$, then $n\in A^k$ and $k\in u^\star(n)$. Since $u^\star(n)$ is only updated when $n\mod 4= 0$, for every round $n\in M$ there exists a round $n'(n) \in r(n)$ such that $t_k(n'(n)+1)-t_k(n'(n))= 1$ since at a round $n'$ such that $n'\mod 4=0$, by construction, the algorithm plays greedily. Therefore we have $\EE[|M|] \leq 4\EE[|\cup_{j>L} A^j_1|+|B|]/\delta$.

\noindent\underline{(ii).2.b) Expected cardinality of $K$}:

\noindent\underline{Expected cardinality of $K'$:}
Let $\overline c_k(n), \overline t_k(n)$ be the \emph{success rate} and associated counter of arm $k$, at time $n$, when played in all other rounds except those in $\cup_{j>L} A^j_1$ and $B$. 
Since $\hat c_k(n) \leq (|\cup_{j>L} A^j_1|+|B| + \overline c_k(n) \overline t_k(n))/t_k(n)$ and $\overline t_k(n))/t_k(n) < 1$ we therefore have that for all $n\in N$:
$$
\hat c_k(n) \leq \delta + \overline c_k(n).
$$
We then bound the number of rounds in $K'$ using Lemma \ref{lem:deviation} (as $\EE[\overline c_k(n)]\leq \nu(L|y^k)$ since $\overline c_k(n)$ is only computed from rounds outside $\cup_{j>L}A^j_1$). In Lemma \ref{lem:deviation} we choose $H = \Lambda =\{n\in K: n\mod 4 = 0\}$, $c=1$ and noting that for all $n\in K'$ we have $r(n)\subset K'$, we obtain that $\EE[|K'|]\leq 4\times 2(2+\delta^{-2})$.

From the above, we then have $\hat c_k(n) - \nu(L|y^k)> 2\delta$ can be bounded by $\EE[|M|] + \EE[|K'|] \leq 4\EE[|\cup_{j>L} A^j_1|+|B|]/\delta + 8(2+\delta^{-2})$.

\noindent\underline{Expected cardinality of $K\setminus K'$:} By the definition of $K$ and $K'$ we have $\hat c_k(n) - \nu(L|y^k) \leq 2\delta$ for all $n\in K\setminus K'$. Note that $\nu(L|u^\star)\geq \nu(L|y^k)$ for all $k>L$. Since $k\notin u^\star$ and $k\in u^\star(n)$ we must have that there exists $y\in u^\star$ such that $y\notin u^\star(n)$. Since the algorithm enters the exploration phase every 4 rounds, $y$ will be played in the last slot of $u(n)$ every $4$ rounds with probability $1/(N-L)$ (due to our tie breaking rule) when $d_y(n) \geq \nu(L|u^\star)-\delta$. We proceed to bound the cardinality of the following sets, for all $y\leq L$:
$$
H_y = \{n\in K\setminus K': y\notin u^\star(n)\text{ and }d_y(n)<\nu(L|u^\star)\}
$$
and
\begin{align*}
J_y = \{n\in K\setminus K': &y\notin u^\star(n)\text{ and } d_y(n)\geq \nu(L|u^\star)\\&\text{ and }\hat c_y(n)\leq \nu(L|u^\star)-\delta\}
\end{align*}

\noindent\underline{Expected cardinality of $H_y$:}
Define $H_y'=\{n\in \NN: d_y(n)< \nu(L|u^\star)\}$ and noting that $H_y\subset H_y'$, we proceed to bound $\EE[|H_y'|]$. We previously bounded the cardinality of $\cup_{j>L} A^j_1$ and hence we have that for all $n\notin \cup_{j>L} A^j_1$:
\begin{equation}\label{eq:correctExpectation}
\EE[ t_y(n+1) \hat c_y(n+1) - t_y(n)\hat c_y(n)]\geq \nu(L|u^\star).
\end{equation}

Denote by $s = |\cup_{j>L} A^j_1|+|B|$. Define $G_y = \{n\in \NN: t_y(n) < s/\delta\}$ and hence for all $n\notin G_y$ we have $\EE[\hat c_y(n)]\geq \nu^\star(L|u^\star) - \delta$. 
We compute the expected number of rounds when $d_y(n) < \nu(L|u^\star)-\delta$ as:
\begin{align*}
\sum\limits_{n=1}^T \PP[d_y(n)< \nu(L|u^\star) ] \leq& \sum\limits_{n\in G_y} \PP[d_y(n) < \nu(L|u^\star) - \delta] + \\&\sum\limits_{n\notin G_y} \PP[d_y(n) < \nu(L|u^\star)-\delta].
\end{align*}
Note that there a.s. $\forall n_0\in G_y$ and $\exists n'<\infty$ such that $\forall n>n'$ we have $s/\delta I(\hat c_y(n_0), \nu(L|u^\star) - \delta) < f(n)$ since $f(n)\to \infty$ when $n\to\infty$ and the left hand side of the inequality is constant. Hence either $|G_y|<\infty$ or $\exists n_1<\infty$ such that for all $n\in G_y$, $n>n_1$ we have $\PP[d_y(n) < \nu(L|u^\star) - \delta]=0$ (i.e. if $y$ is not explored infinitely many times, there almost surely exists a finite time $n_1$ after which for all rounds $n>n_1$ its index $d_y(n)$ is always above $\nu(L|u^\star)$). Consequently, we have that there exists $C_2<\infty$ such that:
\begin{equation}\label{eq:firstPart}
\sum\limits_{n\in G_y} \PP[d_y(n) < \nu(L|u^\star) - \delta]  < C_2
\end{equation}
For every $n\notin G_y$, consider a set of $t_y(n)$ i.i.d. random variables $X^n_i,\ i=1,\dots, t_y(n)$ drawn from a Bernoulli distribution of mean $\gamma = \nu(L|u^\star)-\delta$, and denote by $\hat\gamma(n) = 1/t_y(n) \sum_{i=1}^{t_y(n)} X^n_i$ their observed average. Then from the concentration inequality in Theorem 1 of \cite{garivier2013informational} we have that for all $n$:
$$
\PP[t_y(n) I(\hat\gamma(n), \gamma) > f(n)] \leq  2e \lceil f(n)\log(n)\rceil \exp(-f(n))
$$
and since for $n\notin G_y$ we have $\EE[\hat c_y(n)]\geq \nu(L|u^\star) - \delta$ we then have that:
\begin{align*}
\PP[t_y(n) &I(\hat c_y(n), \gamma) > f(n), \hat c_y(n) < \gamma] \\&\leq \PP[t_y(n) I(\hat \gamma(n), \gamma) > f(n), \hat \gamma(n) < \gamma] \\&\leq  2e \lceil f(n)\log(n)\rceil \exp(-f(n))
\end{align*}
and hence:
\begin{align*}
\sum\limits_{n\notin G_y}& \PP[d_y(n) < \nu(L|u^\star)-\delta] =\\&=\sum\limits_{n\notin G_y} \PP[t_y(n) I(\hat c_y(n), \gamma) > f(n) ,\hat c_y(n) < \gamma]\\&\leq \sum\limits_{n\notin G_y}  2e \lceil f(n)\log(n)\rceil \exp(-f(n)) < \infty
\end{align*}

\noindent\underline{Expected cardinality of $J_y$:}
Now we bound the number of rounds when $d_y(n)> \nu(L|u^\star)-\delta> \hat c_k(n)$ and $k\in u^\star(n)$. In this case, w.p. at least $1/(N-L)$ the algorithm explores $y$ in the last slot of $u(n)$ every $4$ rounds. Note that for all $n\notin \cup_{j>L} A^j_1$ we have:
$$
\EE[t_y(n+1)\hat c_y(n+1) - t_y(n) \hat c_y(n)] \geq \nu(L|u^\star), \forall y\leq L.
$$
Consequently, similar to bounding $\EE[|M|+|K'|]$, we have:
\begin{align*}
\EE[|\{n\notin \cup_{j>L} A^j_1:&  d_y(n)> \hat c_k(n)\text{ and }\hat c_y(n) \leq \nu(L|u^\star)-\delta\}|]\\ \leq& 4(N-L)\EE[|\cup_{j>L} A^j_1|+|B|]/\delta +\\& 8(N-L)(2(N-L)+\delta^{-2})    
\end{align*}

Putting everything together we obtain that there exists $C<\infty$ such that:
\begin{align*}
\EE[|A^k|] \leq &\EE[|A^k_1|]+ 4(N-L+1)\EE[|\cup_{j>L} A^j_1|+|B|]/\delta + \\&8(N-L)(2(N-L)+\delta^{-2})+8(2+\delta^{-2})+C.
\end{align*}

\noindent\underline{(iii) Upper bound on $\EE[C^k]$:} We recall $C^k=\{n\notin \cup_{j>L} A^j\cup B| k\in u(n)$ and $d_k(n)>d_L(n)\}$. From the above we can bound the expected cardinality of the set $\cup_{k>L} A^k_1$. Since for all $n\notin \cup_{k>L}A^k_1$ we then have the expected success rate at round $n$ is:
$$
\EE[t_k(n+1)\hat c_k(n+1) - t_k(n)\hat c_k(n)] \geq \nu(L|y^k)
$$
Hence, similar to bounding $|K'|$, the number of rounds when $\hat c_k(n) \leq \nu(L|y^k) + \delta$ is bounded by:
\begin{align*}
\EE[|\{n\in C^k&:\hat c_k(n) \geq \nu(L|y^k) + \delta\}|] \\&\leq 4/\delta \EE[|\cup_{j>L} A^j_1|+|B|] + 8(2+\delta^{-2}).
\end{align*}
Then, by the definition of our index $d_k(n)$ we have that:
$$
t_k(n) I(\hat c_k(n), \hat c_L(n)) \leq f(n)
$$
and consequently, since forall $n'\in C^k$, $t_k(n')$ is incremented:
\begin{align*}
|C^k|<&t_k(T) \leq f(T) / I(\nu(L|y^k) + \delta, \nu(L|u^\star)) + \\&4(L+1)/\delta \EE[|\cup_{j>L} A^j_1|+|B|] + 8(L+1)(2+\delta^{-2}).
\end{align*}

\noindent\underline{(iv) Upper bound on $\EE[D^k]$:} We recall $D^k=\{n\notin \cup_{j>L} A^j\cup B | k\in u(n)$ and $b_k(n)\geq b_{u^\star_l(n)}(n)$ for some $l$ such that $h(u^\star_l(n))=h(k)\}$. Since an exploration phase and a shuffling of the leader occur alternatively, using Lemma \ref{lem:deviation} we can bound the number of rounds $n\in D^k$ when $|\hat\theta_k(n) - \phi(h(k))\theta_{kh(k)}|>\delta$ by $8(2+\delta^{-2})$ and when $|\hat\theta_l(n) - \phi(h(l))\theta_{lh(l)}|>\delta$ for some $l\leq L$ by $8L(2L+\delta^{-2})$ . Hence, as above, we obtain: 
\begin{align*}
\EE[|D^k|] \leq &f(T)/I(\phi_{h(k)} \theta_{k h(k)}+\delta, \phi_{h(k)} \theta_{L_k h(k)}-\delta)+\\&8(2+\delta^{-2})+8L^2(2L+\delta^{-2}).
\end{align*}
Putting everything together we obtain that for all $n\in\NN$ and $\delta\in (0, \Delta)$ exists $C_1,C_2<\infty$ such that:

\begin{align*}
\EE&\left[\sum_{n\leq T}\mathbbm{1}[k\in u(n)] \right]\leq \frac{f(n) }{ I(\nu(L|y^k) + \delta, \nu(L|u^\star))}  \\ &+\frac{f(n)}{I(\phi_{h(k)} \theta_{k h(k)}+\delta, \phi_{h(k)} \theta_{L_k h(k)}-\delta)} \\ &+[4N(N+2)\delta^{-1}+1][8L(2L+\delta^{-2})\\ & +8L(N-L)(2(N-L)+\delta^{-2})] \\ & +(8L+24)(2+\delta^{-2})+8(N-L)(2(N-L)+\delta^{-2})\\ &+8L^2 (2L+\delta^{-2}) + (4N(N+2)+1)C_1+ C_2.
\end{align*}

Which proves the first statement of the theorem. Furthermore since $u^\star(n) = u^\star$ for all $n\in D^k\cup C^k$ we have that the regret of our algorithm satisfies:
\begin{align*}
\lim\sup\limits_{T\to\infty} \frac{R(T)}{f(T)} &\leq \sum\limits_{k>L} \frac{\mu(u^\star) - \mu(y^k)}{ I(\nu(L|y^k) + \delta, \nu(L|u^\star))} \\&+ \frac{\mu(u^\star) - \mu(w^k)}{I(\phi_{h(k)} \theta_{k h(k)}+\delta, \phi_{h(k)} \theta_{L_k h(k)}-\delta)}
\end{align*}
which concludes the proof.
\clearpage
\bibliographystyle{unsrt}
\bibliography{RA}
\end{document}